\documentclass{article}

\usepackage{microtype}
\usepackage{graphicx}
\usepackage{subfigure}
\usepackage{booktabs} 
\usepackage{csquotes}
\usepackage{caption}
\usepackage{subcaption}
\usepackage{multirow}
\usepackage[monochrome]{color}

\usepackage{hyperref}

\usepackage[accepted]{icml2024}

\usepackage{amsmath}
\usepackage{amssymb}
\usepackage{mathtools}
\usepackage{amsthm}
\usepackage{breqn}
\usepackage{bm}

\usepackage[capitalize,noabbrev]{cleveref}

\theoremstyle{plain}

\theoremstyle{definition}

\theoremstyle{remark}

\usepackage[textsize=tiny]{todonotes}

\icmltitlerunning{RNO: Learning the Smooth Dependence of Solution of PDEs on Geometric Deformations}

\begin{document}

\twocolumn[
\icmltitle{Reference Neural Operators: 
Learning the Smooth Dependence of Solutions of PDEs on Geometric Deformations
           }



\icmlsetsymbol{equal}{*}

\begin{icmlauthorlist}
\icmlauthor{Ze Cheng}{comp}
\icmlauthor{Zhongkai Hao}{yyy}
\icmlauthor{Xiaoqiang Wang}{comp}
\icmlauthor{Jianing Huang}{comp}
\icmlauthor{Youjia Wu}{comp}
\icmlauthor{Xudan Liu}{comp}
\icmlauthor{Yiru Zhao}{comp}
\icmlauthor{Songming Liu}{yyy}
\icmlauthor{Hang Su}{yyy}
\end{icmlauthorlist}

\icmlaffiliation{yyy}{Dept. of Comp. Sci. \& Techn., Institute for AI, BNRist Center, Tsinghua-Bosch Joint ML Center, Tsinghua University}
\icmlaffiliation{comp}{Bosch (China) Investment Co., Ltd.}

\icmlcorrespondingauthor{Hang Su}{suhangss@mail.tsinghua.edu.cn}

\icmlkeywords{Machine Learning, ICML}

\vskip 0.3in
]



\printAffiliationsAndNotice{}  

\begin{abstract}
For partial differential equations on domains of arbitrary shapes, existing works of neural operators attempt to learn a mapping from geometries to solutions. It often requires a large dataset of geometry-solution pairs in order to obtain a sufficiently accurate neural operator. However, for many industrial applications, e.g., engineering design optimization, it can be prohibitive to satisfy the requirement since even a single simulation may take hours or days of computation.
To address this issue, we propose  \textit{reference neural operators} (RNO), a novel way of implementing neural operators, i.e., to learn the smooth dependence of solutions on geometric deformations. Specifically, given a reference solution, RNO can predict solutions corresponding to arbitrary deformations of the referred geometry. This approach turns out to be much more data efficient.
Through extensive experiments, we show that RNO can learn the dependence across various types and different numbers of geometry objects with relatively small datasets. RNO outperforms baseline models in accuracy by a large lead and achieves up to 80\% error reduction.
\end{abstract}

\section{Introduction}


\begin{figure}[t]
    \centering
    \includegraphics[width=0.9\linewidth]{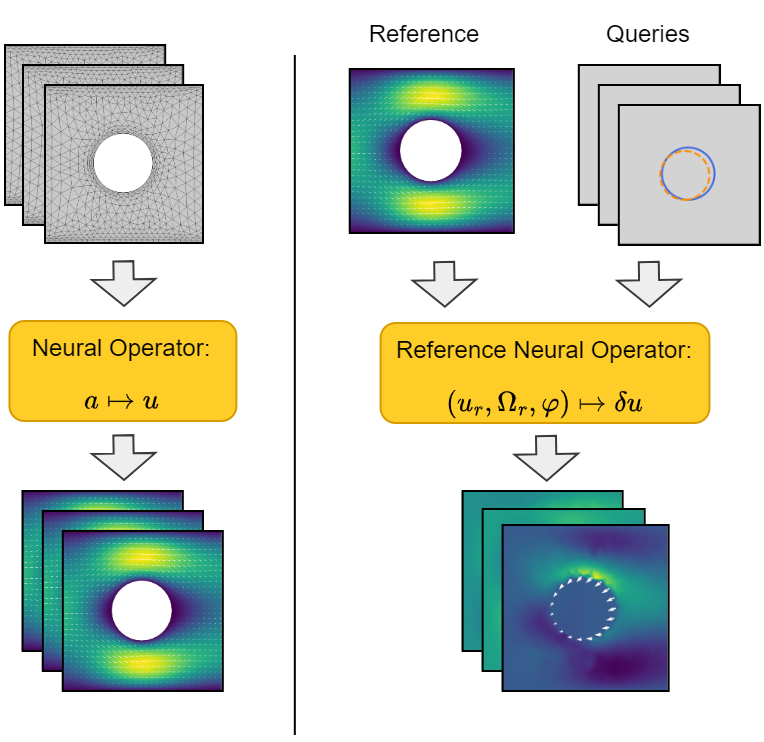}
    \caption{Comparison between two approaches. (Left) Neural operators map directly from geometries/functions $a$ to solutions $u$, which often requires large amount of data to cover various geometries. (Right) Alternatively, given a reference solution $u_r$ on $\Omega_r$, we hope to query solutions to various deformations of its geometry. Let $\varphi: \Omega_r\mapsto\Omega_q$ be a smooth deformation, a reference neural operator can predict the difference between the queried solution and the pushforward of the reference solution $\delta u = u_q - u_r\circ \varphi^{-1}$.}
    \label{fig:NO_vs_RNO}
\end{figure}

Recently, neural operators \cite{lu2021learning, li2020fourier, bhattacharya2021model, nelsen2021random, patel2021physics} as surrogate models for solving partial differential equations (PDEs) have rapidly gained attention due to the success in many applications in physics simulation, e.g., weather forecasting, fluid dynamics, etc. \cite{wang2023scientific, bi2023accurate, zhang2023skilful, pathak2022fourcastnet}. One of the greatest advantages of neural operators is fast inference speed, which may reduce computational cost by orders of magnitude while maintaining good accuracy. 
These methods deal with problems on \textit{fixed} domains, e.g., Earth. However, for engineering design, e.g., sensitivity analysis, optimization and robustness study, iterations of solving PDEs on \textit{deformable} domains are required.
Existing works \cite{shukla2023deep, li2022fourier, li2023geometry, hao2023gnot} of neural operators on deformable domains attempt to learn a mapping from \textcolor{blue}{\textit{geometry space to solution space} (\textbf{G-S})} of PDEs. Such methods require large amount of training data to achieve satisfying accuracy. Unlike some physics simulation problems in fixed domains, such as weather forecasting where climate data is available from research of meteorology, engineering design usually deals with individual cases and needs to simulate various designs to collect data. 
For industrial applications, it often takes hours or days to simulate a single design.
Meanwhile, the geometric design space can be huge, since geometric objects may be a composition of any number of intersection, union and difference. 
Therefore, learning the mapping from geometries to solutions becomes an extremely challenging and expensive task.

To address this problem, we think out of the box and ask a different question: 
\begin{displayquote}
    \textit{With a limited budget on training set, can neural operators learn the change of solution corresponding to the change of geometry?}
\end{displayquote}
See Fig. \ref{fig:NO_vs_RNO} for illustration of the idea. Instead of predicting the solution of an arbitrary design as \textbf{G-S}, we ask, given a reference design what impact some geometric deformation will cause to the reference solution.
Intuitively, similar changes of different geometry shapes can cause similar impact on the solution. For example, considering a flow through a channel with holes, to enlarge, shrink or shift a hole close to the inlet of the channel will cause similar effect to the flow, regardless of how many holes the channel have. 
In this paper, we propose a novel approach, \textit{reference neural operators} (RNO) to predict the change of solution due to the change of geometry. \textbf{G-S} essentially interpolates sparse data in a vast space, and RNO focuses on estimation in the neighborhood of a reference solution. Based on extensive experiments, compared to \textbf{G-S}, RNO turns out to be more data efficient and exhibits superior learning efficacy.


Specifically, RNO takes a reference solution, a corresponding geometry and a deformation to a queried geometry as inputs and outputs its prediction on the difference between reference solution and query solution. 
In the study of shape optimization, the target difference is defined as material derivative \cite{sokolowski1992introduction}. 
It brings several challenges to us, e.g., how to describe a deformation from reference to query? How to combine multiple input functions on irregular meshes in neural network?
Our main contributions include:
\begin{itemize}
    \item We proposed RNO, a novel type of neural operators on irregular meshes. Given a reference solution $(u_r, \Omega_r)$, RNO can estimate the solution of PDEs on deformations of $\Omega_r$.
    \item We provide a simple method of constructing a domain deformation $\varphi$ based on the boundary of two domains.
    \item To handle multiple input functions and strengthen the signal from geometric deformation, we implemented distance-aware cross attention (DACA) in RNO. 
    Since cross attention is known for quadratic complexity and may suffer from scalability issue, additionally, we propose an alternative linear-complexity DACA. To our best knowledge, this is a novel design of linear attention weighted by distance. 
    \item We comprehensively benchmarked RNO in multiple challenging 2D and 3D PDE problems. Results show that even with small size datasets, RNO can learn the smooth dependence of solutions on geometric deformations and outperforms several baseline neural operators.
\end{itemize}

\section{Related Works}

\textbf{Neural operators}.
Learning solution operators of PDEs using neural networks has gained growing attention. A revolutionary wave starts with DeepONet \cite{lu2021learning} and its followup works \cite{wang2021learning, wang2022improved, jin2022mionet}. DeepONet is a meshless model, and it can learn various maps from all sorts of functions, e.g., initial data, boundary conditions, coefficient functions, etc. to solutions. 

Another principled architecture for neural operators is proposed by \cite{li2020neural}, and Fourier Neural Operator (FNO) \cite{li2020fourier} emerged as a successful special case. One of the key ideas is to formulate neural operators as kernel integrals. Similarly, \cite{cao2021choose} reinterpreted attention mechanism as kernel integrals and proposed to construct neural operators by transformers. 
However, all these works are not dedicated to dealing with varying shapes and geometries. In fact, FNO and its variations need to work on regular mesh grids due to the embedded Fast Fourier Transform in FNO layers. 

Geo-FNO \cite{li2022fourier} is designed to handle varying shapes and it learns the operator map from geometries to solutions by a composition of deformation and FNO. The deformation map between regular domains to irregular ones should be either provided in advance or learned from data. GINO \cite{li2023geometry} combines graph neural network with FNO and deals with varying geometry problems in 3D space. NUNO \cite{liu2023nuno} generalizes FNO to irregular meshes by partitioning irregular domains into regular subdomains. GCNN \cite{chen2021graph} learns laminar flow around random 2D objects with graph convolutional neural networks. These works aim to enable generalization of neural operators on shapes and geometries, but such generalization heavily relies on the diversity and the quantity of training samples. For practical applications, where return on investment must be considered, these approaches may be too expensive to apply. 

Similarly, \cite{li2022transformer,hao2023gnot} improved transformer-structured neural operators both on the versatility of input functions and on the performance. While transformer-based models can naturally handle irregular meshes, they still require enough data to learn the mapping from geometries to solutions.

In contrast to all previous methods, RNO aims to learn the change of solution caused by geometric deformation.


\textbf{Hybrid methods}. There exists a type of hybrid methods that combines ML with traditional numerical solvers. In contrast to pure neural operators that only forward pass once, these methods often require iterations. Neural networks work as a replacement of some expensive parts of iterations in numerical solvers. Some representative works include
\cite{kochkov2021machine}, an equation-specific method that replaces an expensive component inside computational fluid dynamics (CFD).
\cite{tompson2017accelerating} and \cite{obiols2020cfdnet} can generalize even to unseen geometries. They also requires being combined with CFD solvers, and the idea is to leverage ML models to accelerate convergence of iterations.
\cite{kahana2023geometry} implements a more subtle way of interweaving ML models with iterative numerical solvers and can generalize to unseen geometries through transfer learning.

However, all these hybrid methods are either problem specific or require tailored combination with numerical solvers. In contrast, RNO does not involve direct interactions with numerical solvers. For unseen geometries, it can refer to a numerically solved example and predict on various deformations.

\textbf{Distance-aware transformers}. Encoding position information in transformers is a vibrant topic \cite{dufter2022position}. For example, \cite{wu2020transformer} proposed distance-aware transformer to re-scale attention weights according to distance between tokens. \textcolor{blue}{OFormer \cite{li2022transformer} and FactFormer \cite{li2024scalable} utilize RoPE \cite{su2024roformer} to encode token position.  These works do not utilize physical distance, e.g., Eulidean distance. On the other hand,} \cite{ying2106transformers} successfully brought transformer to graph-structure problem by encoding structure information including physical distance between nodes. \cite{zhang2023rethinking} theoretically proved that distance is a key ingredient to the expressive power of transformer based GNN models. 

\begin{figure}[t]
    \centering
    \includegraphics[width=1\linewidth]{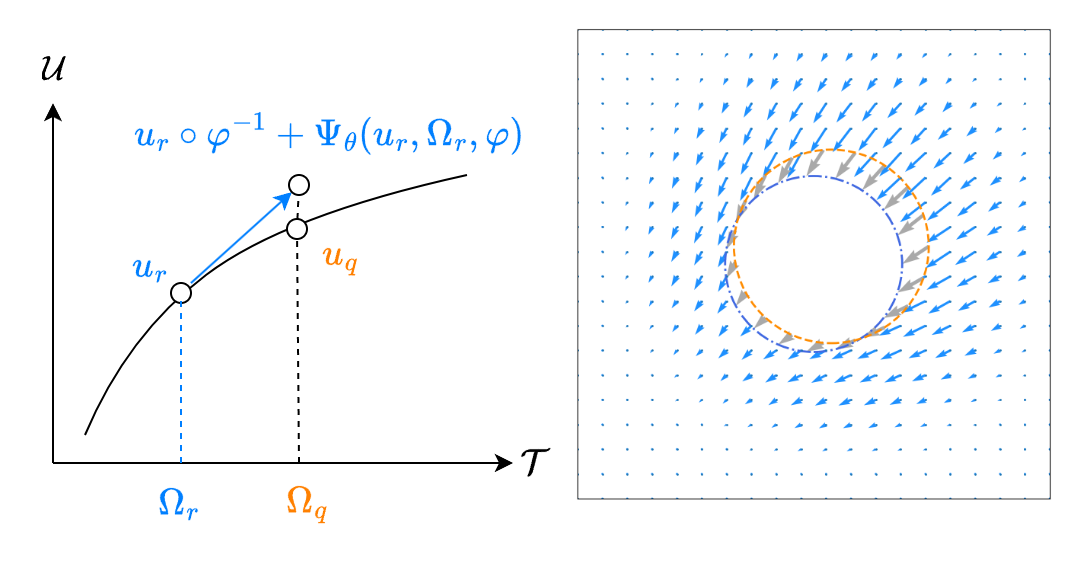}
    \caption{(Left) Given geometry space $\mathcal{T}$ and solution space $\mathcal{U}$, a reference neural operator $\Psi_{\theta}$ learns the material derivative of a solution operator $\mathcal{G}:\mathcal{T}\rightarrow\mathcal{U}$ at a given reference geometry $\Omega_r$. It provides an approximation of the solution on a deformed geometry $\Omega_q=\varphi(\Omega_r)$ with $u_r\circ\varphi^{-1}+\Psi_{\theta}(u_r, \Omega_r, \varphi)$. (Right) The transform (gray vectors) from $\partial\Omega_q$ (orange circle) to $\partial\Omega_r$ (blue circle) is used to construct a vector field that represents a deformation $\varphi^{-1}: \Omega_q \mapsto \Omega_r$.}
    \label{fig:RNO}
\end{figure}

\begin{figure*}[ht]
    \centering
    \includegraphics[width=0.9\linewidth]{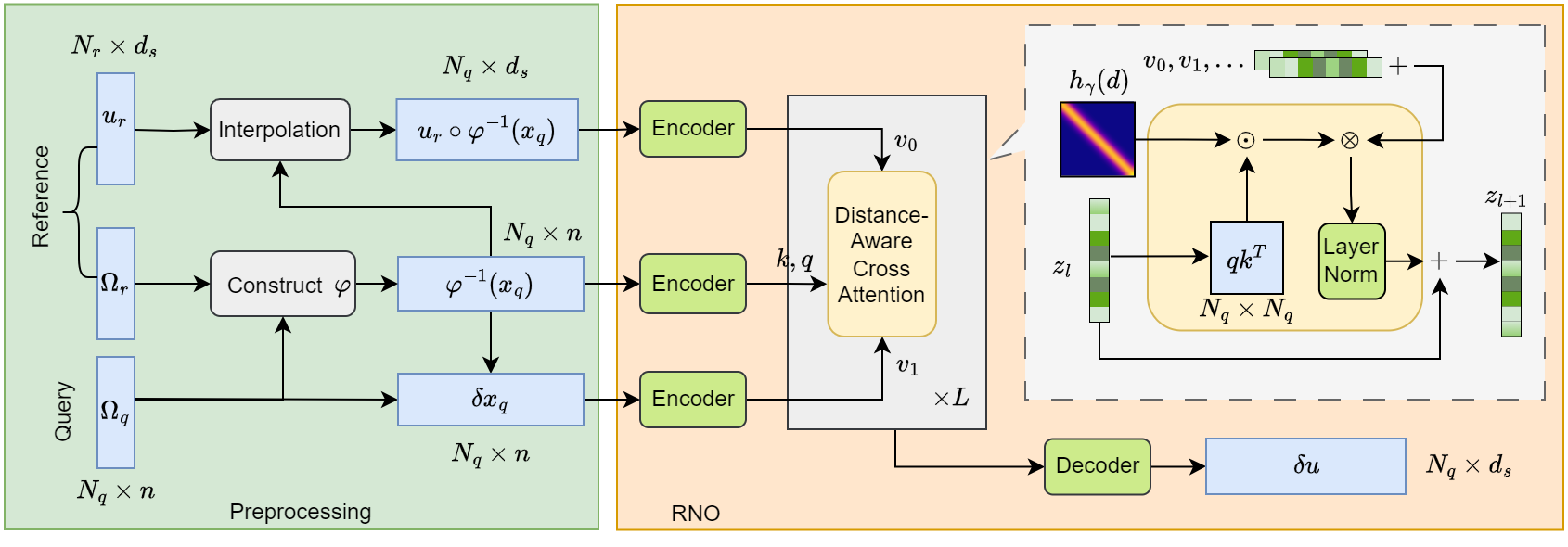}
    \caption{Overview of model architecture which composes of two stages. The first stage is the preprocessing of input data. Input sequences $u_r$, $\Omega_r=\{x_{r_j}\}_j$ and $\Omega_q=\{x_{q_i}\}_i$ are tensors with shape $N_r\times d_s$, $N_r\times n$, $N_q\times n$, \textcolor{blue}{where $n, d_s$ are spatial dimension and dimension of target space}. $\delta x_q = \varphi^{-1}(x_q)-x_q$ is the shift of every point $x_q\in \Omega_q$. The second stage is the forward passing of neural network. $\odot$ is element-wise product, $\otimes$ is matrix product, $+$ is element-wise sum. $\delta u$ is the predicted change of solution, and $\hat{u}_q=u_r\circ\varphi^{-1} + \delta u$.
   }
    \label{fig:arch}
\end{figure*}

\section{Method}
\label{others}

\subsection{Problem Formulation}

Let $D\subset \mathbb{R}^n$ be a Lipschitz domain and $T\in\mathcal{T}$ be a signed distance function such that $\Omega_T=\{x\in D | T(x)>0\}$. Depending on the problem we study, $\Omega_T$ can be either a domain with holes or an interior domain of $D$. Hereafter, $\Omega_T$ can stand for a domain, and when it is self-evident $\Omega_T$ can also stand for an element in the function space $\mathcal{T}$ to simplify notation. Consider the following PDE problem with Dirichlet boundary condition: 
\begin{align}
    \mathcal{N}(u) &= 0 \text{ in } \Omega_T \\
    u &= 0 \text{ on } \partial\Omega_T
\end{align}
where $\mathcal{N}$ is a differential operator and $\partial\Omega_T$ is the boundary of the domain.
Suppose the problem has a unique solution $u_T\in\mathcal{U}_T$, and $\mathcal{U}_T$ is a Banach space of functions on $\Omega_T$. Also, let $\mathcal{U}$ be a Banach space of functions on $D$, and assume $\mathcal{U}_T\subseteq \mathcal{U}$ with a family of linear and bounded extension from $\mathcal{U}_T$ to $\mathcal{U}$. 
Then we can define a solution operator
\begin{equation}
\begin{aligned}
G: \mathcal{T} &\rightarrow \mathcal{U}, \\
 T &\mapsto u_T,
\end{aligned}
\end{equation}
Such operators have some nice properties such as smooth dependence on $T$ within a small neighborhood in $\mathcal{T}$, given some suitable conditions. For elliptic, parabolic and hyperbolic problems, readers can find classic analysis in \cite{sokolowski1992introduction}. For Stokes equation and Navier-Stokes equation, shape holomorphy of solutions is analyzed by \cite{cohen2018shape}.

Given a reference solution $(u_r, \Omega_r)$, we hope to learn about solution $u_q$ on $\Omega_q$, where $\Omega_q$ is a small perturbation of $\Omega_r$, i.e., there exists a smooth deformation $\varphi$ with smooth inverse, i.e., $\varphi, \varphi^{-1}\in C^k(D;D)$, $k\geq 1$, such that 
\begin{equation}
    \varphi:\Omega_r\mapsto \Omega_q.
\end{equation}
See Fig. \ref{fig:RNO} (Left) for conceptual illustration, \textcolor{blue}{and the construction of $\varphi$ will be discussed in Section \ref{algo}.} Formally, we define $\Psi_{\theta}$ as a neural operator parameterized by $\theta$,
\begin{equation}
    \begin{aligned}
    \Psi_{\theta}: \mathcal{U}\times\mathcal{T}\times C^k(\mathcal{T}) &\rightarrow\mathcal{U} \\
    (u_r, \Omega_r, \varphi) &\mapsto u_q - u_r\circ \varphi^{-1}, 
\end{aligned}
\end{equation}
and its training objective is
\begin{align}\label{objective0}
    \min_{\theta}\mathbb{E}[\|\Psi_{\theta}(u_r, \Omega_r, \varphi) - (u_q - u_r\circ \varphi^{-1})\|],
\end{align}
where $\|\cdot\|$ is the norm on $\mathcal{U}$. In practice, the norm usually is chosen to be $L^p$-norm, $p>0$, for computational simplicity. 
In Appendix \ref{app:model_details}, we derive this objective from material derivative.

\textcolor{red}{Suppose every queried domain $\Omega_q$ has a set of points $\{x_{q_i}\}_i\subset\Omega_q$ $i=1,\cdots,N_q$. For $N$ queries, each one is paired with a reference triplet $(u_r, \Omega_r, \varphi)$.} Then the objective function \eqref{objective0} is approximated by
\begin{dmath}\label{objective}
    \min_{\theta}  \frac{1}{N}\sum_q\sum_i \|\Psi_{\theta}(u_r, \Omega_r, \varphi)(x_{q_i}) - (u_q - u_r\circ\varphi^{-1})(x_{q_i})\|.
\end{dmath}
Let $\delta u = \Psi_{\theta}(u_r, \Omega_r, \varphi)$ 
and then predicted solution $\hat{u}_q = u_r\circ\varphi^{-1} + \delta u$. Given ground truth $u_q$, the above objective is implemented as the metric loss between $\hat{u}_q$ and $u_q$.


\textcolor{blue}{Note that we assume the deformation $\varphi$ and its inverse $\varphi^{-1}$ between reference and query domains exist and are smooth. It sets the boundary for application of the proposed method. Namely, we should not apply the method to cases where no such deformation exist and expect good learning results. For example, a query has different topology from a reference by introducing or removing a hole. In fact, such topological change can cause sharp change to solutions.}

\subsection{Reference Neural Operator}

Following the principle for neural operators established by \cite{li2020neural, li2020fourier}, we also formulate the architecture of RNO as a stack of kernel integrals,
\begin{align}
    \Psi = \mathcal{Q} \circ \mathcal{L}_L \circ \cdots \circ \mathcal{L}_1\circ \mathcal{P}.
\end{align}
$\mathcal{P}$ is an encoder that lifts function values to hidden space \textcolor{blue}{$\mathbb{R}^s$, where $s$ is feature dimension,} and it should be able to encode different numbers of geometries. 
$\mathcal{L}_l$, $l=1,\cdots,L$ are integral operator layers.
$\mathcal{Q}: \mathbb{R}^s \rightarrow \mathbb{R}^{d_s}$ is a decoder that projects hidden variables back \textcolor{blue}{to target space $\mathbb{R}^{d_s}$}.

\subsubsection{Geometry Encoder} 
If the domain $\Omega_T$ we consider have multiple geometry objects, i.e., the boundary of $\Omega_T$ has multiple components, $\partial\Omega_T=\cup_i \Gamma_i \cup \partial D$ where $\partial D$ is the boundary of $D$ (or simply $\partial\Omega_T=\cup_i \Gamma_i$ if $\Omega_T$ is an interior domain), we need to design an encoder capable of handling these different numbers of components.
Inspired by \cite{molinaro2023neural}, we encode each change of geometries by a shared encoder $\mathcal{P}: \mathcal{U}\times\mathcal{T}\times C^k(\mathcal{T})\rightarrow \mathbb{R}^{s}$ and sum their outputs, where $s$ is the dimension of features. Suppose at a point $x$ with value $u_r(x)$, $\mathcal{P}$ is defined as

\begin{small}
    \begin{align}
    v(x) = \mathcal{P}(x, u_r(x), \Omega_r, \varphi) = \sum_i \mathcal{P} (x, u_r(x), \Gamma_i, \varphi).
\end{align}
\end{small}
Unlike \cite{molinaro2023neural} where output of $\mathcal{P}$ is averaged, here we sum all outputs to reflect the composition of changes of all geometry components.
For any location $x\in\Omega_q$, we encode it to $\mathcal{P}(x)$ in the latent space $\mathbb{R}^s$. Hereafter, when it is self-evident, we still use $x$ to refer latent encoding of these locations for simplicity. 

Here for the encoder, we use the geometric parameters to represent $\Gamma_i$'s, e.g., centers, radius, length, etc. 
Thus, we simply takes the difference between geometric parameters $p_r,p_q$ of reference and query to represent $\varphi$.
Note that we also represent $\varphi$ as a vector function on mesh grids and combine this information by cross attention. See next sections for cross attention and construction of vector function $\varphi$.

\subsubsection{Distance-Aware Cross-Attention}\label{DACA}
In order to handle multiple input functions, e.g., reference solution and deformation, let's consider an integral operator $\mathcal{K}:\mathcal{U}\times\cdots\times\mathcal{U}\rightarrow\mathcal{U}$ with $M$ input functions $v^j$, $j=1,\cdots,M$,
\begin{small}
    \begin{align}\label{int_oper}
    w(x) = \mathcal{K}(v^1, \cdots, v^M)(x) = \int_D \sum_{j=1}^M\kappa_j(x, y)v^j(y) dy
\end{align}
\end{small}
Note that we are dealing with irregular meshes deformable geometries. Inspired by \cite{cao2021choose, hao2023gnot}, we apply attention mechanism \cite{vaswani2017attention} to approximate the kernels $\kappa_j(\cdot, \cdot)$ on irregular meshes. 
For $\bm{q}, \bm{k}, \bm{v}\in\mathbb{R}^s$ and a sequence of inputs $X=\{x_i\}_{1\leq i\leq N}$,, we have $\bm{q}(X), \bm{k}(X), \bm{v}^j(X)\in\mathbb{R}^{N\times s}$. Then attention works as a kernel and is defined by 
\begin{align}
    \textbf{attn}(x, y_i) = \text{softmax}(\bm{q}(x) \bm{k}^T(y_i)).
\end{align}
Distance weight is helpful to strengthen attention according to spatial relation between elements of $\bm{q}$ and $\bm{k}$ \cite{ying2106transformers, zhang2023rethinking}. Considering the change of solution can be strongly related to the location of deformation in some problems, e.g., fluid dynamics, we apply distance weighting and 
approximate \eqref{int_oper} by a distance-aware cross attention (DACA) layer,
\begin{equation}\label{crossAT}
    \bm{w}(x) \approx \textcolor{blue}{\frac{1}{\alpha}} \sum^N_{i=1} \sum^M_{j=1} \textbf{attn}_j(x, y_i) \cdot h_{\gamma}(d(x, y_i)) \cdot \bm{v}^j(y_i) ,
\end{equation}
where $d$ is a distance function, e.g., Euclidean distance, $h(t) = e^{-\frac{t^2}{\gamma^2}}$ and $\gamma$ is a hyperparameter. \textcolor{blue}{$\alpha=\sum_i\sum_j \textbf{attn}_j(x, y_i) \cdot h_{\gamma}(d(x, y_i))$ is a normalization factor.} Each $\textbf{attn}_j$ can have its own $\bm{q}^j$ and $\bm{k}^j$ to learn different kernels $\kappa_j(\cdot, \cdot)$, but the computation cost would be $O(MN^2)$ for each layer. In practice, we find a shared kernel in each layer sufficient for our purpose and keep the cost as $O(N^2)$. In Appendix \ref{app:linearAT}, we describe a DACA layer of linear complexity.

Finally, we construct each integral operator layer $\mathcal{L}_l$ as
\begin{equation}
        \bm{v}_{l+1}(x) = \mathcal{L}_l(\bm{v}_l)(x) = \bm{v}_l(x) + f\left( \bm{w}_l(x)\right), 
\end{equation}
where $\bm{w}_l$ is defined by \eqref{crossAT} and $f$ is a composition of a layer normalization and an MLP with nonlinear activation function \textcolor{blue}{GELU \cite{hendrycks2016gaussian}}. For input functions $v^j$ of $\bm{w}_l$, we choose the triplet $\left(\bm{v}_l, \mathcal{P}_{1}(u_r\circ\varphi^{-1}(x_q)), \mathcal{P}_2(\varphi^{-1}(x_q)-x_q)\right)$, where $\mathcal{P}_i$'s, $i=1,2$, are encoders that lift input function values to hidden variables. See overall architecture in Fig. \ref{fig:arch}.

\begin{table*}[ht]
    \centering
    \begin{tabular}{cc|cc|ccc|c}
                 Dataset & Component  & GNOT & Geo-FNO & R-GNOT & R-FNO-i & R-MIONet  & RNO (Ours)   \\
         \hline
        \hline
        \multirow{3}{7em}{NS2d-360} 
                                    & $u$ & 1.6e-1  & 4.6e-1 & 1.0e-1 & 2.7e-1 & 3.7e-1 & \textbf{3.9e-2}    \\
                                     & $v$ & 4.2e-1  & 9.0e-1 & 2.5e-1 & 7.5e-1 & 6.6e-1 & \textbf{1.1e-1}    \\
                                     & $p$ & 2.1e-1  & 4.3e-1 & 9.5e-2 & 2.1e-1 & 3.6e-1 & \textbf{4.2e-2}    \\
        \hline
        \multirow{3}{7em}{NS2d-sq-360} 
                                        & $u$ & 1.1e-1  & 3.4e-1 & 8.6e-2 & 2.3e-1 & 3.2e-1 & \textbf{5.3e-2}    \\
                                        & $v$ & 3.2e-1 & 8.5e-1 & 2.7e-1 & 7.3e-1 & 7.0e-1 & \textbf{1.8e-1}    \\
                                        & $p$ & 2.0e-1  & 4.3e-1 & 1.6e-1 & 2.4e-1 & 4.4e-1 & \textbf{9.1e-2}    \\
        \hline
        \multirow{3}{7em}{Inductor2d-320}
                                            & $A_z$ & 5.8e-2  &  1.4e-1 & 2.5e-2 & 3.0e-2 & 3.8e-2 & \textbf{4.3e-3 }  \\
                                           & $B_x$ & 6.2e-2  &  3.4e-1 & 4.5e-2 & 5.5e-2 & 1.1e-1 & \textbf{1.7e-2}   \\
                                           & $B_y$ & 7.0e-2  &  4.2e-1 & 5.6e-2 & 6.0e-2 & 1.2e-1 & \textbf{2.2e-2}   \\
        \hline
        \multirow{4}{7em}{Heatsink3d-80} 
                                        & $u$ & 1.1e-1 & - & 5.0e-2 & - & 2.0e-1 & \textbf{4.2e-2}   \\
                                       & $v$ & 3.2e-1 & - & 1.9e-1 & - & 5.6e-1 & \textbf{1.6e-1}   \\
                                       & $w$ & 2.6e-1 & - & 1.4e-1* & - & 4.8e-1 & \textbf{1.4e-1}   \\
                                       & $T$ & 8.4e-3 & - & 4.3e-3 & - & 1.9e-2 & \textbf{3.7e-3}   \\
    \end{tabular}
    \caption{The number following the name of datasets is the size of datasets. The prefix ``R-'' indicates modified models that take reference solution $u_r$ and deformation $\varphi$ as inputs. See definition of baseline models in Section \ref{experiment}. \textbf{Bold} is the best. *The 2nd digit after the decimal is higher than RNO.}
    \label{tab:main_results}
\end{table*}

\begin{algorithm}[t]
   \caption{Forward function of RNO}
   \label{alg:RNO}
\begin{algorithmic}
   \STATE {\bfseries Input:} Query $\left(x_q, p_q, \Gamma_q \right)$ and reference $\left((x_r, u_r), p_r, \Gamma_r \right)$  
   
   {\color{lightgray} \# Preprocessing}
   \textcolor{blue}{\STATE $\delta x_q \leftarrow Construct\_Phi(\Gamma_q, \Gamma_r, x_q)$}
   \STATE $x_s \leftarrow x_q + \delta x_q$
   \STATE $u_{interp} \leftarrow Interpolate((x_r, u_r), x_s)$ 
   
   \quad
   
   {\color{lightgray} \# Forward passing of neural network}
   \STATE $\hat{u}_q \leftarrow u_{interp} + \Psi_{\theta}(x_q, u_{interp}, \delta x_q, p_q, p_r)$
    \STATE {\bfseries Output:} $\hat{u}_q$

\end{algorithmic}
\end{algorithm}

\subsection{Algorithms}\label{algo}


In this section, we deal with the following questions. How to construct a vector function to represent deformation $\varphi$? How to obtain a pushforward function $u_r\circ\varphi^{-1}$?

\textbf{Interpolation for pushforward}. Notice that the pushforward of $u_r$, $u_r\circ\varphi^{-1}(x_q)$, can be obtained by interpolation. One can construct a triangular mesh on $\Omega_r$ from reference data $\{(x_{r_i}, u_{r_i})\}_i$, and then interpolate $\{u_r\circ\varphi^{-1}(x_{q_j})\}_j$.

\textcolor{red}{\textbf{Ref-Query Dataloader}}. We need a dataloader that loads data in pairs of reference and query. If training dataset is constructed in pairs, where two solutions and geometries in one pair are close to each other, reference and query are paired naturally. If training set is constructed by sampling from geometric design space, for each query, reference can be determined by K-nearest-neighbor. Here the distance between examples can be Euclidean distance between geometric parameters.

\textcolor{blue}{In our implementation}, a dataset has format $\left((x, u), p, \Gamma \right)$, where $p$ stands for the geometry parameters and $\Gamma$ is the set of boundary points. Then a paired data loaded from the dataloader has format $\left((x_q, u_q), (p_q, p_r), ((x_r, u_r), \Gamma_r, \Gamma_q )\right)$.

\textbf{Constructing $\varphi$}. $\varphi$ should avoid folding or tearing which means positive Jacobian everywhere. 
In practice, we take a na\"ive approach as approximation.
\textbf{Step 1.} Find shifting vectors from points on boundaries of $\Omega_r$ to points on boundaries of $\Omega_q$. This requires 1-to-1 matching between points. If the deformation from reference to query is obtained by some deformation methods, e.g., free-form deformation \cite{sederberg1986free}, then the 1-to-1 matching is automatic. Or we can reconstruct boundaries of simple geometries with parameters to create such matching.
\textbf{Step 2.} For each point $x_q$, set its shifting vector equal to the shifting vector of the closest boundary point, and then put on the weight by the distance between $x_q$ and the closest boundary point. The weight function is $h(t) = e^{-\frac{t^2}{\gamma_{\varphi}^2}}$, and $\gamma_{\varphi}$ is another hyperparameter.
\textbf{Step 3.} Protect the wall $\partial D$ of domain $D$, i.e., to prevent points being shifted outside of the domain. Let $d$ be the distance between $x_r$ and $\partial D$. We define a smooth cutoff function on $[0, \infty)$,
\begin{equation}
    \eta(d) = \left\{
    \begin{array}{ll}
        0,  &d=0 \\
        e^{1-\frac{d_{max}^2}{d^2}},  &0<d<d_{max} \\
        1,  &d\geq d_{max}
    \end{array} \right.
\end{equation}
where $d_{max}$ is chosen to be the shortest distance between $\Gamma_i$'s to $\partial D$. Then apply the cutoff function on all shifting vectors obtained from step 2. Note that all steps are smooth procedures except step 1. Here, our choice of step 1 may not be smooth, but it provides a simple yet effective approximation of smooth $\varphi$. See Fig. \ref{fig:RNO} (Right) for the illustration of the construction.

\textbf{Preprocessing and RNO}. Given a dataset, all previous procedures are processed before feeding data into forward pass of neural networks. The pipeline is summarized in Algorithm \ref{alg:RNO}. Obviously, for scaling purpose, one may move the entire procedure of preparing reference and query pairs outside of the training loop of RNO.

\section{Experiments}\label{experiment}

\textbf{Datasets}. We consider several PDE problems in both 2D and 3D space. Problems are defined on various geometric domains, including different shapes, different positions, different sizes and different numbers of geometric objects. All datasets are generated in pairs. For each pair, the two domains are deformations of each other, and during training and testing the two data samples are used as reference and query for each other. We intentionally limit the size of datasets in order to evaluate all methods for practical use. The dataset sizes of 2D problems are no more than 400, and for more expensive 3D problems, we use a dataset of 80 samples. Training and testing sets are split in ratio $8:2$. For more details of each dataset, please check Appendix \ref{app:dataset}. For all problems, we choose $\gamma_{\varphi}=0.1$ to construct deformation $\varphi$, see Section \ref{algo}. 
\begin{itemize}
    \item \textbf{Steady 2D-Navier-stokes equations} models flows though a 2D channel with holes. There are two types of holes, circle (\textbf{NS2d}) and square (\textbf{NS2d-sq}). The number of holes ranges from $1\sim 9$, and the size and the position of holes varies.
    \item \textbf{Maxwell equations (Inductor2d)} is a model of circular shape inductor with different numbers of circular copper wire.
    \item \textbf{Heat transfer and CFD (Heatsink3d)} is a 3D model of heatsink with hexagon shaped pin-fins. The number of pin-fins ranges from $2\sim 14$. The gap between rows of pin-fins varies.
\end{itemize}

\textcolor{blue}{The above challenging datasets are featured with random number of simple geometries which can be deformed by shifting and resizing. In Appendix \ref{app:add_exp}, we benchmarked RNO from a different angle, namely to expand example scenarios to complex geometry shapes such as airfoil. With two more datasets, Airfoil-Euler\cite{li2022fourier} and Airfoil-RANS\cite{bonnet2022airfrans}, we showcase that RNO can flexibly be applied to free-form type of deformations beyond previous examples.} 

\begin{figure*}[!ht]
    \centering
    \includegraphics[width=0.9\linewidth]{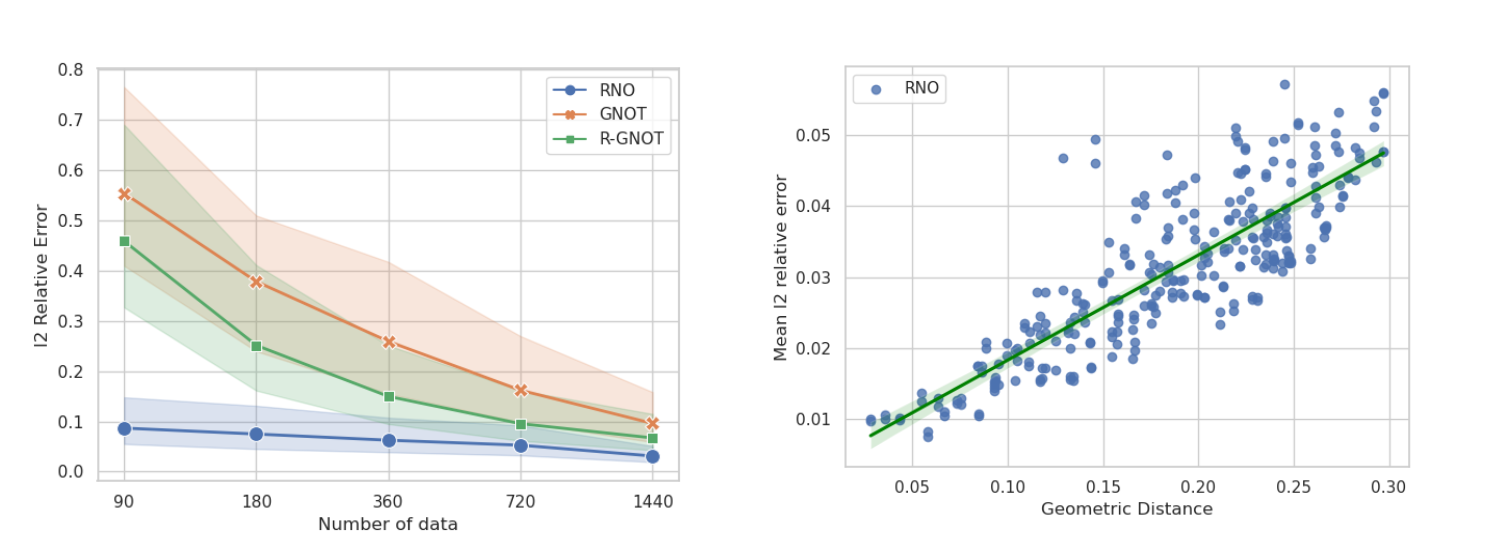}
    \caption{(Left) Mean error of all components versus the size of dataset. (Right) Error versus distance.}
    \label{fig:scaling}
\end{figure*}

\textbf{Baselines}. Also, we compare RNO to several baseline models. In particular, we modify some of the models to learn the operator: $(u_r, \Omega_r, \varphi)\mapsto u_q$. The modified models are prefixed by ``R-''.
\begin{itemize}
    \item \textbf{MIONet} \cite{jin2022mionet} is similar to DeepONet \cite{lu2021learning} as a mesh-free model, and it requires fixed sensors in the domain. So we interpolate all input data on uniform grids as sensors. 
    \item \textbf{Geo-FNO} \cite{li2023geometry} is able to learn deformations of the domain and generalize on different geometric shapes. It can work with irregular meshes, so we make no change on this model \textcolor{blue}{except for downsampling data with the same number of grids}.
    \item \textbf{FNO} \cite{li2020fourier} should be applied on uniform grids. So we interpolate data on 64 by 64 uniform grids. Additionally, we modify the input of FNO and name it as R-FNO-i.
    \item \textbf{GNOT} \cite{hao2023gnot} is versatile with the formats of input functions, including geometric parameters, boundary shapes. etc. So, we use vanilla GNOT as one of the baseline models. Additionally, we also use R-GNOT as a baseline.
\end{itemize}
All baselines except MIONet use up to 4 times more parameters than RNO. MIONet uses roughly 30\% fewer parameters than RNO.


\textbf{Evaluation metric and hyperparameters}. The metric of evaluation is $l_2$ relative error. Suppose $u$ and $\hat{u}$ are ground truth solution and predicted solution,
\begin{equation}
    \|u-\hat{u}\|_{rel_2} =  \left(\frac{\sum_{i}|u_i - \hat{u}_i|^2}{\sum_i |u_i|^2}\right)^{\frac{1}{2}}.
\end{equation}
We train all models with AdamW optimizer \cite{loshchilov2017decoupled} with cyclical learning rate \cite{smith2017cyclical}. All experiments are ran 100 epochs with batchsize $1\sim 4$ depending on the number of nodes in meshes. 

\subsection{Main Results}


In Table \ref{tab:main_results}, we summarize main results of the experiment. There are two types of baseline models, with or without reference solution $(u_r, \Omega_r)$ and discretized deformation $\varphi$ as inputs, distinguished by prefix ``R-''. Models without ``R-'' belong to traditional \textbf{G-S} neural operators.
RNO outperforms all baseline models in all problems.

Note that R-GNOT outperforms GNOT for all problems, indicating that the extra information about reference solution and deformation is helpful for prediction. Meanwhile, R-GNOT is still not as good as RNO, suggesting that the performance of RNO is not only due to extra inputs but a key fact that the target of RNO is approximating the material derivative of solution operators.

The performance of linear RNO (RNO-L) is summarized in Table \ref{tab:linear_RNO}.
RNO-L has a little higher error most likely due to the randomness introduced by RFM. We point out that a drawback of linear DACA is $O(d^2D)$ memory cost. When $D$ is taken a high integer value to improve the accuracy of estimation of the distance weight function $h_{\gamma}(d)$, the memory cost grows by $D$ times compared to vanilla linear attention. However, it can be mitigated by increasing number of heads of attention, with number of heads $n_{head}$, the memory cost is reduced to $O(\frac{d^2D}{n_{head}})$.

A natural baseline of RNO may be the pushforward of reference solution, $u_r\circ\varphi^{-1}$, i.e., $\Psi_{\theta}\equiv 0$. We compare the mean $l_2$ relative error of the pushforward reference solution and RNO. See Table \ref{tab:pushforward} in Appendix \ref{app:error},  RNO has smaller errors for all problems, and some error is reduced by $50\%\sim 80\%$. It shows that RNO has essentially learned to predict the difference $u_q-u_r\circ\varphi^{-1}$.

\begin{figure*}[h]
    \centering
    \includegraphics[width=0.99\linewidth]{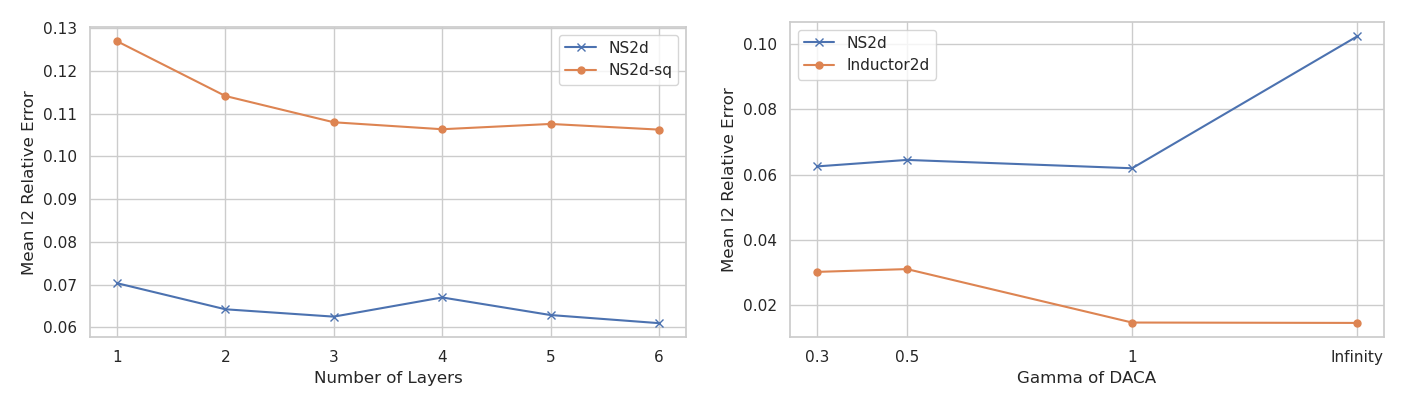}
    \caption{(Left) Error versus number layers. (Right) Error versus $\gamma$ of DACA.}
    \label{fig:layer_gamma}
\end{figure*}

\subsection{Scaling Experiment}
\textbf{Size of dataset}. In the discussion above, we see that even with small dataset, RNO is able to learn the change of solution due to the change of geometry and hence provides an approximation of the queried solution. We wonder how size of dataset would affect the performance of RNO. We choose the dataset NS2d which has 1440 samples in total. Then we down sample the dataset to smaller datasets with sampling rate $r=1,2,4,8,16$. The baseline models are GNOT and R-GNOT, which are two best models besides RNO. In Figure \ref{fig:scaling} (Left), we plot GNOT and R-GNOT representing vanilla and modified neural operators against RNO on NS2d. The error on three models all decrease as size of dataset increases. Though with more data the gap between RNO and the other methods narrows, RNO consistently outperforms the other two. We emphasize that the ability of learning from small amount of data is highly desirable to shape optimization and other engineering design problem.

Notice that some components are more challenging than others to predict, and $l_2$ relative errors can be higher than 10\%, see Table \ref{tab:main_results}. However, this problem can be mitigated when the size of dataset increases. On the full NS2d dataset with 1440 samples, GNOT and R-GNOT still have an error larger than 10\% on $v$, but RNO achieved reducing error by almost 50\%. See Table \ref{tab:scaling}.

\textbf{Number layers of RNO}. The performance of RNO will gain little improvement for number of layers $L>3$, see Fig. \ref{fig:layer_gamma} (Left), so we choose $L=3$ to balance cost and benefit.

\begin{table}[t]
    \centering
    \begin{tabular}{c|ccc}
        \hline
         & $u$ & $v$ & $p$ \\
         \hline
        GNOT & 5.9e-2 & 1.6e-1 & 7.1e-2\\
        R-GNOT & 4.3e-2 & 1.1e-1 & 4.4e-2 \\
        RNO & \textbf{1.9e-2} & \textbf{5.1e-2} & \textbf{2.3e-2} \\
        \hline
    \end{tabular}
    \caption{Results of models on NS2d-1440.}
    \label{tab:scaling}
\end{table}

\subsection{Error Analysis}
Since RNO learns to approximate the material derivative of solution operators, a natural question is how geometric distance relates to prediction error In Figure \ref{fig:scaling} (Right), we plot the mean $l_2$ relative error of components versus the geometric distance for NS2d, and error grows with distance. Here the geometric distance is defined by Euclidean distance between geometry parameters of reference and query. 
In general, for non-parametric geometries, we can define the norm of discretized deformation as the geometric distance. See more error analysis for other datasets in Appendix \ref{app:error}.

\subsection{The Hyperparameter of DACA}
The hyperparameter $\gamma$ in \eqref{crossAT} is crucial to the performance of RNO, and it is problem dependent. See Fig. \ref{fig:layer_gamma} (Right). For NS2d, the mean $l_2$ relative error increases as $\gamma$ increases. One way of reasoning this effect is that for fluid dynamics, a local geometry change will most likely impact nearby flow. On the other hand, for Inductor2d, the mean $l_2$ relative error decreases as $\gamma$ increases. This interesting opposite trend is due to the well-known fact that magnetic-electronic phenomenon is governed by Maxwell equations, and such elliptic equations have slow-decay kernel functions for solutions \cite{evans2022partial}. Therefore, long distance  impact will be significant.
In our implementation, for NS2d, NS2d-sq and Heatsink3d, we choose $\gamma=0.3$. For Inductor2d, we choose $\gamma=\infty$ (disable distance weighting).

\subsection{Ablation Study}
\begin{table}[t]
    \centering
    \begin{tabular}{c|ccc}
        \hline
                    & NS2d & NS2d-sq & Inductor2d \\
         \hline
        w/o $\delta u$ &  7.4e-2 & 1.2e-1 & 3.6e-2 \\
        w/o DACA & 9.9e-2 & 2e-1 & 3.8e-2\\
        All & \textbf{6.3e-2} & \textbf{1.1e-1} & \textbf{1.5e-2}\\
        \hline
    \end{tabular}
    \caption{Ablation study on learning target and DACA.}
    \label{tab:ablation}
\end{table}

We conduct ablation study on the components of RNO.
Results in Table \ref{tab:ablation}  show that both the target $\delta u$ and DACA are key components. Values are mean $l_2$ relative errors. ``Without DACA'' means to remove entire DACA layers.``Without $\delta u$'' means to learn the operator $(u_r, \Omega_r, \varphi)\mapsto u_q$ instead of $(u_r, \Omega_r, \varphi)\mapsto \delta u$, which is the same as ``R-'' type of baseline models except for model architecture.  
To rationalize why predicting $\delta u$ is better than predicting $u_q$ directly, RNO explicitly determines the pushforward $u_r\circ\varphi^{-1}$ in $\hat{u}$, which reduces the complexity of the target.

\section{Conclusion}
Learning the operator mapping from geometry shapes to solutions of PDEs is a challenging problem due to the fact that geometry space is extremely huge and complicated. It can be prohibitive to sample sufficient data from geometry space and train neural operators. We propose an alternative approach that neural operators learn the change of a reference solution corresponding to a queried deformation. We conduct extensive experiments and show that RNO can learn this target efficiently. Our method may direct a new and practical path of applying neural operators as surrogate models for shape optimization and other downstream tasks.

\section*{Acknowledgements}
This work was supported by the National Key Research and Development Program of China (No. 2020AAA0106302), NSFC Projects (Nos.~92370124, 62350080, 92248303, 62276149). We thank Prof. Genggeng Huang of Fudan University for helpful discussion and suggestion. We also thank the reviews for their comments from which we deeply benefits. Finally, we wish to recognize the collaboration between Bosch and Tsinghua University, which provides the opportunity for this joint research.

\section*{Impact Statement}
This paper presents work whose goal is to advance the field of Machine Learning as new tools for physics simulation. There are many potential societal consequences of our work, none which we feel must be specifically highlighted here.

\bibliography{RNO}

\begin{thebibliography}{45}
\providecommand{\natexlab}[1]{#1}
\providecommand{\url}[1]{\texttt{#1}}
\expandafter\ifx\csname urlstyle\endcsname\relax
  \providecommand{\doi}[1]{doi: #1}\else
  \providecommand{\doi}{doi: \begingroup \urlstyle{rm}\Url}\fi

\bibitem[Bhattacharya et~al.(2021)Bhattacharya, Hosseini, Kovachki, and Stuart]{bhattacharya2021model}
Bhattacharya, K., Hosseini, B., Kovachki, N.~B., and Stuart, A.~M.
\newblock Model reduction and neural networks for parametric pdes.
\newblock \emph{The SMAI journal of computational mathematics}, 7:\penalty0 121--157, 2021.

\bibitem[Bi et~al.(2023)Bi, Xie, Zhang, Chen, Gu, and Tian]{bi2023accurate}
Bi, K., Xie, L., Zhang, H., Chen, X., Gu, X., and Tian, Q.
\newblock Accurate medium-range global weather forecasting with 3d neural networks.
\newblock \emph{Nature}, 619\penalty0 (7970):\penalty0 533--538, 2023.

\bibitem[Bonnet et~al.(2022)Bonnet, Mazari, Cinnella, and Gallinari]{bonnet2022airfrans}
Bonnet, F., Mazari, J., Cinnella, P., and Gallinari, P.
\newblock Airfrans: High fidelity computational fluid dynamics dataset for approximating reynolds-averaged navier--stokes solutions.
\newblock \emph{Advances in Neural Information Processing Systems}, 35:\penalty0 23463--23478, 2022.

\bibitem[Cao(2021)]{cao2021choose}
Cao, S.
\newblock Choose a transformer: Fourier or galerkin.
\newblock \emph{Advances in neural information processing systems}, 34:\penalty0 24924--24940, 2021.

\bibitem[Chen et~al.(2021)Chen, Hachem, and Viquerat]{chen2021graph}
Chen, J., Hachem, E., and Viquerat, J.
\newblock Graph neural networks for laminar flow prediction around random two-dimensional shapes.
\newblock \emph{Physics of Fluids}, 33\penalty0 (12), 2021.

\bibitem[Cohen et~al.(2018)Cohen, Schwab, and Zech]{cohen2018shape}
Cohen, A., Schwab, C., and Zech, J.
\newblock Shape holomorphy of the stationary navier--stokes equations.
\newblock \emph{SIAM Journal on Mathematical Analysis}, 50\penalty0 (2):\penalty0 1720--1752, 2018.

\bibitem[Dufter et~al.(2022)Dufter, Schmitt, and Sch{\"u}tze]{dufter2022position}
Dufter, P., Schmitt, M., and Sch{\"u}tze, H.
\newblock Position information in transformers: An overview.
\newblock \emph{Computational Linguistics}, 48\penalty0 (3):\penalty0 733--763, 2022.

\bibitem[Evans(2022)]{evans2022partial}
Evans, L.~C.
\newblock \emph{Partial differential equations}, volume~19.
\newblock American Mathematical Society, 2022.

\bibitem[Hao et~al.(2023)Hao, Wang, Su, Ying, Dong, Liu, Cheng, Song, and Zhu]{hao2023gnot}
Hao, Z., Wang, Z., Su, H., Ying, C., Dong, Y., Liu, S., Cheng, Z., Song, J., and Zhu, J.
\newblock Gnot: A general neural operator transformer for operator learning.
\newblock In \emph{International Conference on Machine Learning}, pp.\  12556--12569. PMLR, 2023.

\bibitem[Hendrycks \& Gimpel(2016)Hendrycks and Gimpel]{hendrycks2016gaussian}
Hendrycks, D. and Gimpel, K.
\newblock Gaussian error linear units (gelus).
\newblock \emph{arXiv preprint arXiv:1606.08415}, 2016.

\bibitem[Jin et~al.(2022)Jin, Meng, and Lu]{jin2022mionet}
Jin, P., Meng, S., and Lu, L.
\newblock Mionet: Learning multiple-input operators via tensor product.
\newblock \emph{SIAM Journal on Scientific Computing}, 44\penalty0 (6):\penalty0 A3490--A3514, 2022.

\bibitem[Kahana et~al.(2023)Kahana, Zhang, Goswami, Karniadakis, Ranade, and Pathak]{kahana2023geometry}
Kahana, A., Zhang, E., Goswami, S., Karniadakis, G., Ranade, R., and Pathak, J.
\newblock On the geometry transferability of the hybrid iterative numerical solver for differential equations.
\newblock \emph{Computational Mechanics}, pp.\  1--14, 2023.

\bibitem[Kochkov et~al.(2021)Kochkov, Smith, Alieva, Wang, Brenner, and Hoyer]{kochkov2021machine}
Kochkov, D., Smith, J.~A., Alieva, A., Wang, Q., Brenner, M.~P., and Hoyer, S.
\newblock Machine learning--accelerated computational fluid dynamics.
\newblock \emph{Proceedings of the National Academy of Sciences}, 118\penalty0 (21):\penalty0 e2101784118, 2021.

\bibitem[Li et~al.(2020{\natexlab{a}})Li, Kovachki, Azizzadenesheli, Liu, Bhattacharya, Stuart, and Anandkumar]{li2020fourier}
Li, Z., Kovachki, N., Azizzadenesheli, K., Liu, B., Bhattacharya, K., Stuart, A., and Anandkumar, A.
\newblock Fourier neural operator for parametric partial differential equations.
\newblock \emph{arXiv preprint arXiv:2010.08895}, 2020{\natexlab{a}}.

\bibitem[Li et~al.(2020{\natexlab{b}})Li, Kovachki, Azizzadenesheli, Liu, Bhattacharya, Stuart, and Anandkumar]{li2020neural}
Li, Z., Kovachki, N., Azizzadenesheli, K., Liu, B., Bhattacharya, K., Stuart, A., and Anandkumar, A.
\newblock Neural operator: Graph kernel network for partial differential equations.
\newblock \emph{arXiv preprint arXiv:2003.03485}, 2020{\natexlab{b}}.

\bibitem[Li et~al.(2022{\natexlab{a}})Li, Huang, Liu, and Anandkumar]{li2022fourier}
Li, Z., Huang, D.~Z., Liu, B., and Anandkumar, A.
\newblock Fourier neural operator with learned deformations for pdes on general geometries.
\newblock \emph{arXiv preprint arXiv:2207.05209}, 2022{\natexlab{a}}.

\bibitem[Li et~al.(2022{\natexlab{b}})Li, Meidani, and Farimani]{li2022transformer}
Li, Z., Meidani, K., and Farimani, A.~B.
\newblock Transformer for partial differential equations' operator learning.
\newblock \emph{arXiv preprint arXiv:2205.13671}, 2022{\natexlab{b}}.

\bibitem[Li et~al.(2023)Li, Kovachki, Choy, Li, Kossaifi, Otta, Nabian, Stadler, Hundt, Azizzadenesheli, et~al.]{li2023geometry}
Li, Z., Kovachki, N.~B., Choy, C., Li, B., Kossaifi, J., Otta, S.~P., Nabian, M.~A., Stadler, M., Hundt, C., Azizzadenesheli, K., et~al.
\newblock Geometry-informed neural operator for large-scale 3d pdes.
\newblock \emph{arXiv preprint arXiv:2309.00583}, 2023.

\bibitem[Li et~al.(2024)Li, Shu, and Barati~Farimani]{li2024scalable}
Li, Z., Shu, D., and Barati~Farimani, A.
\newblock Scalable transformer for pde surrogate modeling.
\newblock \emph{Advances in Neural Information Processing Systems}, 36, 2024.

\bibitem[Liu et~al.(2023)Liu, Hao, Ying, Su, Cheng, and Zhu]{liu2023nuno}
Liu, S., Hao, Z., Ying, C., Su, H., Cheng, Z., and Zhu, J.
\newblock Nuno: A general framework for learning parametric pdes with non-uniform data.
\newblock \emph{arXiv preprint arXiv:2305.18694}, 2023.

\bibitem[Loshchilov \& Hutter(2017)Loshchilov and Hutter]{loshchilov2017decoupled}
Loshchilov, I. and Hutter, F.
\newblock Decoupled weight decay regularization.
\newblock \emph{arXiv preprint arXiv:1711.05101}, 2017.

\bibitem[Lu et~al.(2021)Lu, Jin, Pang, Zhang, and Karniadakis]{lu2021learning}
Lu, L., Jin, P., Pang, G., Zhang, Z., and Karniadakis, G.~E.
\newblock Learning nonlinear operators via deeponet based on the universal approximation theorem of operators.
\newblock \emph{Nature machine intelligence}, 3\penalty0 (3):\penalty0 218--229, 2021.

\bibitem[Molinaro et~al.(2023)Molinaro, Yang, Engquist, and Mishra]{molinaro2023neural}
Molinaro, R., Yang, Y., Engquist, B., and Mishra, S.
\newblock Neural inverse operators for solving pde inverse problems.
\newblock \emph{arXiv preprint arXiv:2301.11167}, 2023.

\bibitem[Multiphysics(1998)]{multiphysics1998introduction}
Multiphysics, C.
\newblock Introduction to comsol multiphysics{\textregistered}.
\newblock \emph{COMSOL Multiphysics, Burlington, MA, accessed Feb}, 9\penalty0 (2018):\penalty0 32, 1998.

\bibitem[Nelsen \& Stuart(2021)Nelsen and Stuart]{nelsen2021random}
Nelsen, N.~H. and Stuart, A.~M.
\newblock The random feature model for input-output maps between banach spaces.
\newblock \emph{SIAM Journal on Scientific Computing}, 43\penalty0 (5):\penalty0 A3212--A3243, 2021.

\bibitem[Obiols-Sales et~al.(2020)Obiols-Sales, Vishnu, Malaya, and Chandramowliswharan]{obiols2020cfdnet}
Obiols-Sales, O., Vishnu, A., Malaya, N., and Chandramowliswharan, A.
\newblock Cfdnet: A deep learning-based accelerator for fluid simulations.
\newblock In \emph{Proceedings of the 34th ACM international conference on supercomputing}, pp.\  1--12, 2020.

\bibitem[Patel et~al.(2021)Patel, Trask, Wood, and Cyr]{patel2021physics}
Patel, R.~G., Trask, N.~A., Wood, M.~A., and Cyr, E.~C.
\newblock A physics-informed operator regression framework for extracting data-driven continuum models.
\newblock \emph{Computer Methods in Applied Mechanics and Engineering}, 373:\penalty0 113500, 2021.

\bibitem[Pathak et~al.(2022)Pathak, Subramanian, Harrington, Raja, Chattopadhyay, Mardani, Kurth, Hall, Li, Azizzadenesheli, et~al.]{pathak2022fourcastnet}
Pathak, J., Subramanian, S., Harrington, P., Raja, S., Chattopadhyay, A., Mardani, M., Kurth, T., Hall, D., Li, Z., Azizzadenesheli, K., et~al.
\newblock Fourcastnet: A global data-driven high-resolution weather model using adaptive fourier neural operators.
\newblock \emph{arXiv preprint arXiv:2202.11214}, 2022.

\bibitem[Peng et~al.(2021)Peng, Pappas, Yogatama, Schwartz, Smith, and Kong]{peng2021random}
Peng, H., Pappas, N., Yogatama, D., Schwartz, R., Smith, N.~A., and Kong, L.
\newblock Random feature attention.
\newblock \emph{arXiv preprint arXiv:2103.02143}, 2021.

\bibitem[Rahimi \& Recht(2007)Rahimi and Recht]{rahimi2007random}
Rahimi, A. and Recht, B.
\newblock Random features for large-scale kernel machines.
\newblock \emph{Advances in neural information processing systems}, 20, 2007.

\bibitem[Sederberg \& Parry(1986)Sederberg and Parry]{sederberg1986free}
Sederberg, T.~W. and Parry, S.~R.
\newblock Free-form deformation of solid geometric models.
\newblock In \emph{Proceedings of the 13th annual conference on Computer graphics and interactive techniques}, pp.\  151--160, 1986.

\bibitem[Serrano et~al.(2024)Serrano, Le~Boudec, Kassa{\"\i}~Koupa{\"\i}, Wang, Yin, Vittaut, and Gallinari]{serrano2024operator}
Serrano, L., Le~Boudec, L., Kassa{\"\i}~Koupa{\"\i}, A., Wang, T.~X., Yin, Y., Vittaut, J.-N., and Gallinari, P.
\newblock Operator learning with neural fields: Tackling pdes on general geometries.
\newblock \emph{Advances in Neural Information Processing Systems}, 36, 2024.

\bibitem[Shukla et~al.(2023)Shukla, Oommen, Peyvan, Penwarden, Bravo, Ghoshal, Kirby, and Karniadakis]{shukla2023deep}
Shukla, K., Oommen, V., Peyvan, A., Penwarden, M., Bravo, L., Ghoshal, A., Kirby, R.~M., and Karniadakis, G.~E.
\newblock Deep neural operators can serve as accurate surrogates for shape optimization: a case study for airfoils.
\newblock \emph{arXiv preprint arXiv:2302.00807}, 2023.

\bibitem[Smith(2017)]{smith2017cyclical}
Smith, L.~N.
\newblock Cyclical learning rates for training neural networks.
\newblock In \emph{2017 IEEE winter conference on applications of computer vision (WACV)}, pp.\  464--472. IEEE, 2017.

\bibitem[Sokolowski \& Zol{\'e}sio(1992)Sokolowski and Zol{\'e}sio]{sokolowski1992introduction}
Sokolowski, J. and Zol{\'e}sio, J.-P.
\newblock \emph{Introduction to shape optimization}.
\newblock Springer, 1992.

\bibitem[Su et~al.(2024)Su, Ahmed, Lu, Pan, Bo, and Liu]{su2024roformer}
Su, J., Ahmed, M., Lu, Y., Pan, S., Bo, W., and Liu, Y.
\newblock Roformer: Enhanced transformer with rotary position embedding.
\newblock \emph{Neurocomputing}, 568:\penalty0 127063, 2024.

\bibitem[Tompson et~al.(2017)Tompson, Schlachter, Sprechmann, and Perlin]{tompson2017accelerating}
Tompson, J., Schlachter, K., Sprechmann, P., and Perlin, K.
\newblock Accelerating eulerian fluid simulation with convolutional networks.
\newblock In \emph{International Conference on Machine Learning}, pp.\  3424--3433. PMLR, 2017.

\bibitem[Vaswani et~al.(2017)Vaswani, Shazeer, Parmar, Uszkoreit, Jones, Gomez, Kaiser, and Polosukhin]{vaswani2017attention}
Vaswani, A., Shazeer, N., Parmar, N., Uszkoreit, J., Jones, L., Gomez, A.~N., Kaiser, {\L}., and Polosukhin, I.
\newblock Attention is all you need.
\newblock \emph{Advances in neural information processing systems}, 30, 2017.

\bibitem[Wang et~al.(2023)Wang, Fu, Du, Gao, Huang, Liu, Chandak, Liu, Van~Katwyk, Deac, et~al.]{wang2023scientific}
Wang, H., Fu, T., Du, Y., Gao, W., Huang, K., Liu, Z., Chandak, P., Liu, S., Van~Katwyk, P., Deac, A., et~al.
\newblock Scientific discovery in the age of artificial intelligence.
\newblock \emph{Nature}, 620\penalty0 (7972):\penalty0 47--60, 2023.

\bibitem[Wang et~al.(2021)Wang, Wang, and Perdikaris]{wang2021learning}
Wang, S., Wang, H., and Perdikaris, P.
\newblock Learning the solution operator of parametric partial differential equations with physics-informed deeponets.
\newblock \emph{Science advances}, 7\penalty0 (40):\penalty0 eabi8605, 2021.

\bibitem[Wang et~al.(2022)Wang, Wang, and Perdikaris]{wang2022improved}
Wang, S., Wang, H., and Perdikaris, P.
\newblock Improved architectures and training algorithms for deep operator networks.
\newblock \emph{Journal of Scientific Computing}, 92\penalty0 (2):\penalty0 35, 2022.

\bibitem[Wu et~al.(2020)Wu, Wu, and Huang]{wu2020transformer}
Wu, C., Wu, F., and Huang, Y.
\newblock Da-transformer: Distance-aware transformer.
\newblock \emph{arXiv preprint arXiv:2010.06925}, 2020.

\bibitem[Ying et~al.(2021)Ying, Cai, Luo, Zheng, Ke, He, Shen, and Liu]{ying2106transformers}
Ying, C., Cai, T., Luo, S., Zheng, S., Ke, G., He, D., Shen, Y., and Liu, T.
\newblock Do transformers really perform bad for graph representation? arxiv 2021.
\newblock \emph{arXiv preprint arXiv:2106.05234}, 2021.

\bibitem[Zhang et~al.(2023{\natexlab{a}})Zhang, Luo, Wang, and He]{zhang2023rethinking}
Zhang, B., Luo, S., Wang, L., and He, D.
\newblock Rethinking the expressive power of gnns via graph biconnectivity.
\newblock \emph{arXiv preprint arXiv:2301.09505}, 2023{\natexlab{a}}.

\bibitem[Zhang et~al.(2023{\natexlab{b}})Zhang, Long, Chen, Xing, Jin, Jordan, and Wang]{zhang2023skilful}
Zhang, Y., Long, M., Chen, K., Xing, L., Jin, R., Jordan, M.~I., and Wang, J.
\newblock Skilful nowcasting of extreme precipitation with nowcastnet.
\newblock \emph{Nature}, 619\penalty0 (7970):\penalty0 526--532, 2023{\natexlab{b}}.

\end{thebibliography}
\bibliographystyle{icml2024}

\newpage
\appendix
\onecolumn
\section{Datasets}\label{app:dataset}
All datasets are generated by  by COMSOL 6.0. All data has format $\left((x, u), p, \Gamma \right)$, where $p$ stands for the geometry parameters and $\Gamma$ is the set of boundary points. Then we construct a dataloader to load a pair of data as $\left((x_q, u_q), (p_q, p_r), ((x_r, u_r), \Gamma_r, \Gamma_q )\right)$. All experiments run on NVIDIA Tesla V100.

\begin{figure}[h]
    \centering
    \includegraphics[width=1\linewidth]{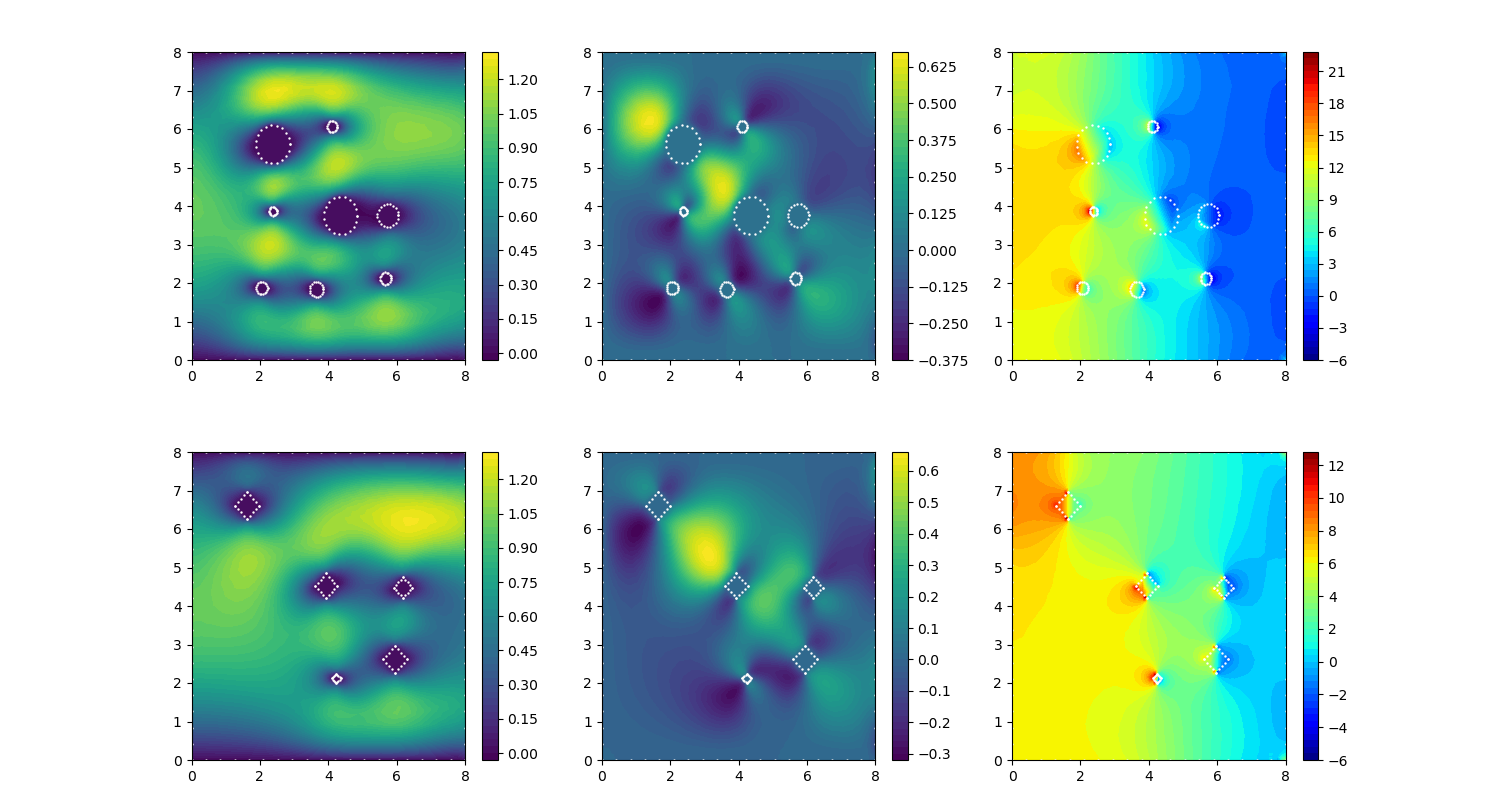}
    \caption{Columns are $u,v,p$. Rows are circle and square. White dots are boundary points.}
    \label{fig:ns2d}
\end{figure}

\textbf{NS2d and NS2d-sq} describes fluid flows through a square channel with holes, namely, $D=[0, 8]\times [0, 8]$, $\Omega=D\setminus\bigcup_{i=1}^M R_i$, where $M=1, \cdots, 9$ and $R_i$'s are disks or squares. Boundary condition is ``no-slip''. Steady Navier-Stokes equation is
\begin{align}
         \bm{u}\cdot\nabla\bm{u} &= \frac{1}{\text{Re}} \nabla^2 \bm{u} - \nabla p   \\
         \nabla\cdot\bm{u} &= 0 
\end{align}
where $\text{Re}=1$ is the Reynolds number and $\bm{u}=(u,v)$. We sample uniformly the number of holes with random positions and sizes. Additionally, for each sample we perturb its geometry parameters and make them a reference and query pair. In total, there are 1440 samples, 1152 for training and 288 for testing. The pairing of data is kept after splitting. For smaller dataset experiments, we down sample the dataset. Below are some examples of the dataset, see Fig. \ref{fig:ns2d}.

\textbf{Inductor2d} models a circular-shape magnetic core wrapped by some number of copper coils. The governing equations are steady Maxwell equations,
\begin{align}
    \nabla \times \bm{H} &= \bm{J} \\
    \bm{B} &= \nabla \times \bm{A} \\
    \bm{J} &= \sigma\bm{E} \\
    \bm{B} &= \mu_0\mu_r\bm{H}
\end{align}
with interface condition
\begin{align}
    \bm{n}\times \bm{A} = 0
\end{align}
where $\mu_0$ is the vacuum permeability and $\mu_r$ is the permeability of the magnetic core.
For 2d problem, suppose $\bm{A} = (0, 0, A_z)$ and $\bm{B}=(\partial_y A_z, -\partial_x A_z, 0)$, the above equations simplify to an elliptic type of equation of $A_z$. Particularly, if $\mu_r$ is a constant, it is a Poisson equation. 
\begin{figure}
    \centering
    \includegraphics[width=\linewidth]{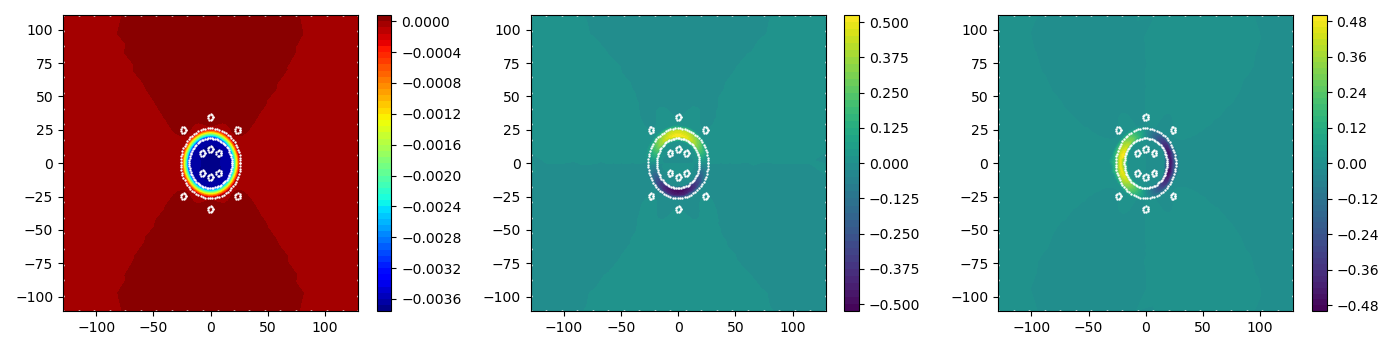}
    \caption{Plots of $A_z, B_x, B_y$. White dots are boundary points.}
    \label{fig:inductor}
\end{figure}

\textbf{Heatsink3d} models a heatsink with pin-fins. The detail of the model is omitted. Readers can find details in \cite{multiphysics1998introduction}.
\begin{figure}[h]
    \centering
    \includegraphics[width=\linewidth]{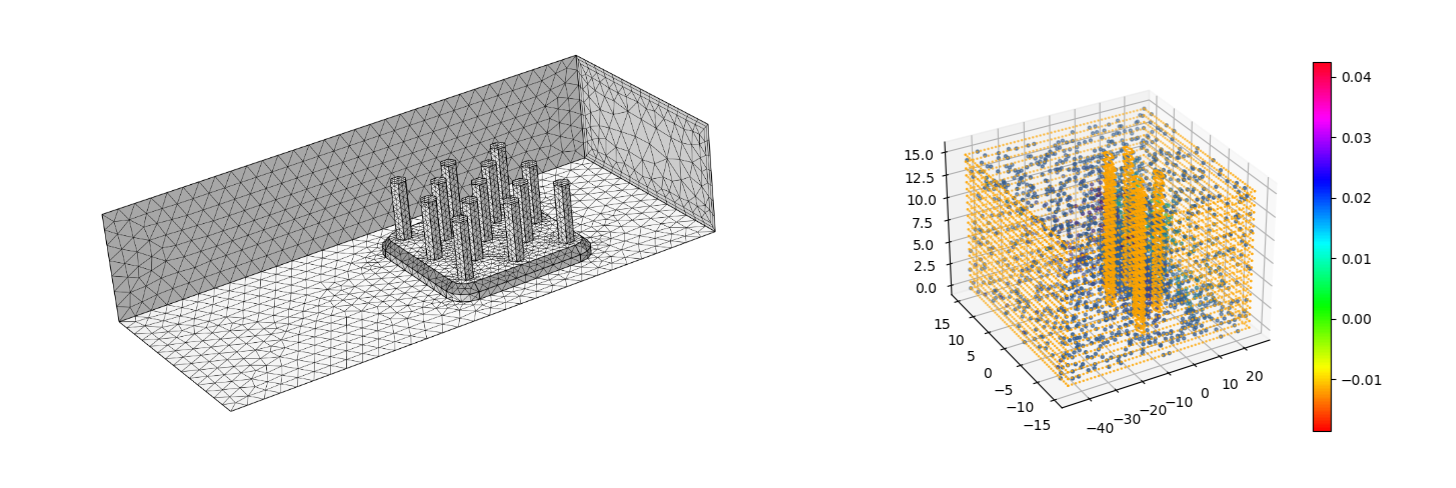}
    \caption{(Left) The mesh of a 3d model of heatsink with 10 hexgon pin-fins. (Right) A data sample of 6 pin-fins. The color of each point in space is the value of a component of $u,v,w,T$. Orange dots are boundary points.}
    \label{fig:heatsink}
\end{figure}

\textcolor{blue}{
\section{Additional Experiments}\label{app:add_exp}
In this section, we present benchmark on two additional datasets, Airfoil-Euler \cite{li2022fourier} and Airfoil-RANS \cite{bonnet2022airfrans} to demonstrate RNO working with complex geometry shapes and free-form type of deformations. The differences between this section and previous experiments are: 1. The number of geometry object is fixed to one, but the type of deformations is extended beyond shifting and resizing to squeezing, stretching, rotation, etc.; 2. When testing, instead of pairing queries and references among testing set, we find the nearest neighbor of a test query from training set and use it as reference. Note that this setting mimics practical usage of RNO in real world. The nearest neighbor is determined by geometric distance, which is implemented as either distance between geometry parameters or norm of deformation.}

\begin{figure}
    \centering
    \includegraphics[width=\linewidth]{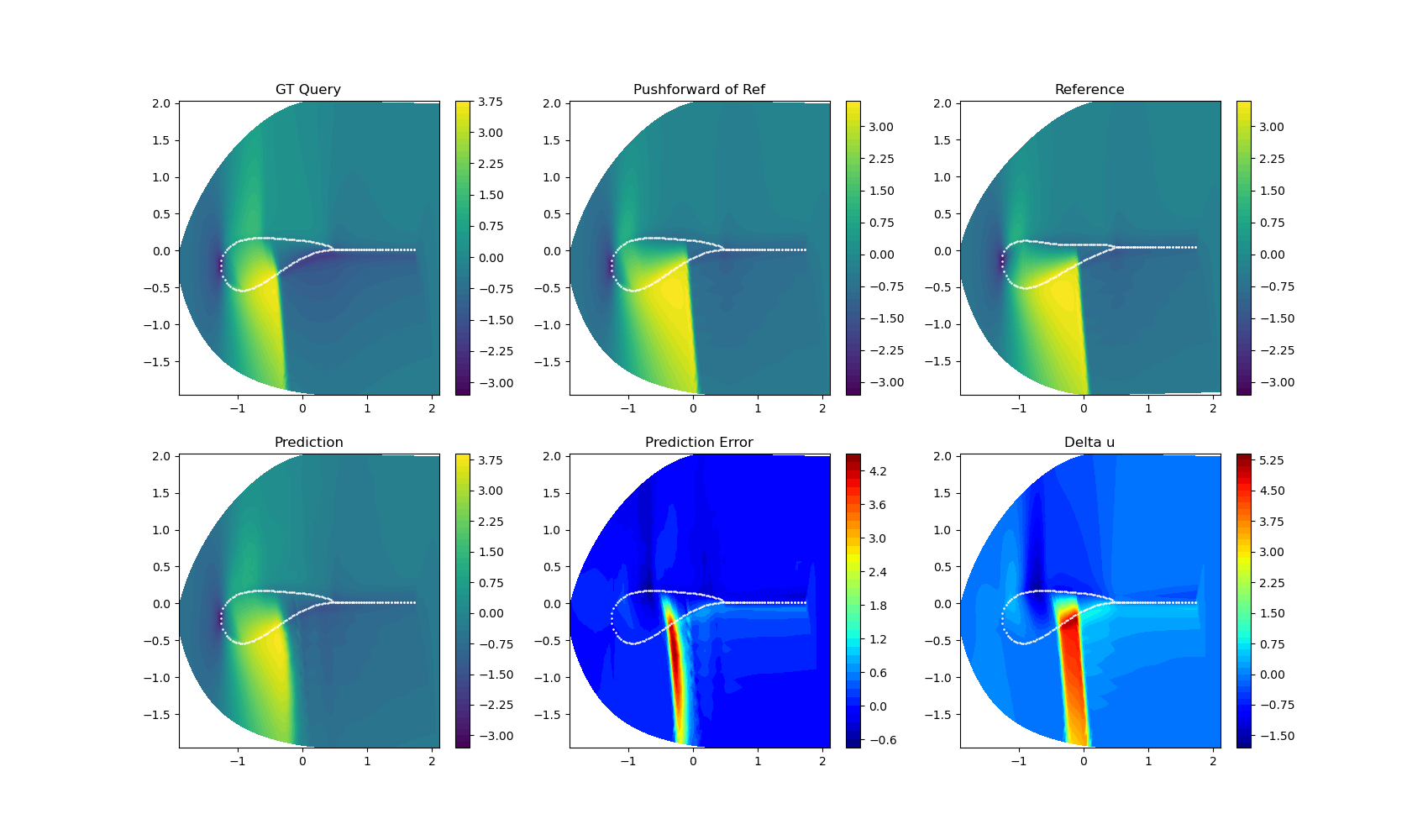}
    \caption{Qualitative example of RNO on the fluid density of Airfoil-Euler.}
    \label{fig:airf_euler}
\end{figure}

\textcolor{blue}{
\textbf{Airfoil-Euler} is a simplified model of airfoil with Euler equation with no viscosity, where the boundary conditions are fixed and also angle of attack is set to zero. The target physical field is fluid density, which is a 1D field. See details in \cite{li2022fourier}. We notice that many state of the art neural operators have achieved quite low error rates, order of magnitude $1e-3\sim 1e-2$ on this dataset, see e.g., \cite{serrano2024operator}. One reason is that airfoil area is less than 1\% of the whole domain, and the error is averaged to be low since most area is almost constant and easy to fit. Another reason is that the training set contains 1000 samples. Therefore, we made two changes to the dataset: 1. Cropping the data to a smaller neighborhood of airfoil (see Fig. \ref{fig:airf_euler}); 2. Limiting training set to 200 samples to fit our setting, which is also called ``scarce data regime" in \cite{bonnet2022airfrans}.
}

\textcolor{blue}{
Another feature of this dataset is that all mesh grids are deformed from one standard uniform mesh grids, so that the tensors of mesh grids have the same shape. As a result, a natural 1-1 mapping of grids exists among data, which saves the need of constructing deformation from reference to query and the following interpolation, greatly simplifying implementation of RNO. Baseline models are CORAL \cite{serrano2024operator}, NU-FNO \cite{liu2023nuno}, Geo-FNO \cite{li2022fourier} and GNOT \cite{hao2023gnot}. RNO outperforms all other baselines on this modified dataset, see Table \ref{tab:airfoil_euler}. More importantly, it shows RNO can be applied to general deformations. Metric is $l_2$, $\|u-\hat{u}\|_{l_2} = \frac{1}{N}\sum_{i}^N|u_i - \hat{u}_i|^2$. 
}

\textcolor{blue}{
\begin{table}[h!]
    \centering
    \begin{tabular}{c|ccccc}
        \hline
        Model & CORAL & NU-FNO & Geo-FNO & GNOT & RNO \\
        \hline
        $l_2$ Error & 8.1e-2 & 1.5e-1 & 7.9e-2 & 4.0e-2 & \textbf{3.4e-2}\\
        \hline
    \end{tabular}
    \caption{$l_2$ Error on Airfoil-Euler.}
    \label{tab:airfoil_euler}
\end{table}
}

\textcolor{blue}{
\textbf{Airfoil-RANS} is a much more challenging CFD dataset generated from simulation of Reynolds-Averaged Navier–Stokes (RANS), which contains complex patterns of turbulent viscosity. Besides varying geometry, the boundary conditions, e.g., inlet velocity, angle of attack, Reynolds number, etc., also vary in the dataset,  It brings challenge to determine suitable reference data for RNO. Ideally, reference and query should have the same boundary conditions and small difference in geometry. However, we manage to show that RNO still outperforms \textbf{G-S} type neural operators with only 180 training data. It shows that the methodology of RNO can be applied to real-life problems where data can be scarce and problems can be complicated.}

\textcolor{blue}{
In our implementation, we use both geometry shape and boundary conditions to determine the top-3 nearest neighbors as reference for training, and use the top-1 neighbor in training set as reference for testing. We found that different target components prefers different strategies of reference. For example, dominant factors for pressure may be inlet velocity and angle of attack. So, a closer reference on these two factor can help predicting pressure. We feel there is great potential to explore, however, due to scope of this paper, we present our primitive study on this dataset.
}

\begin{figure}
    \centering
    \includegraphics[width=1\linewidth]{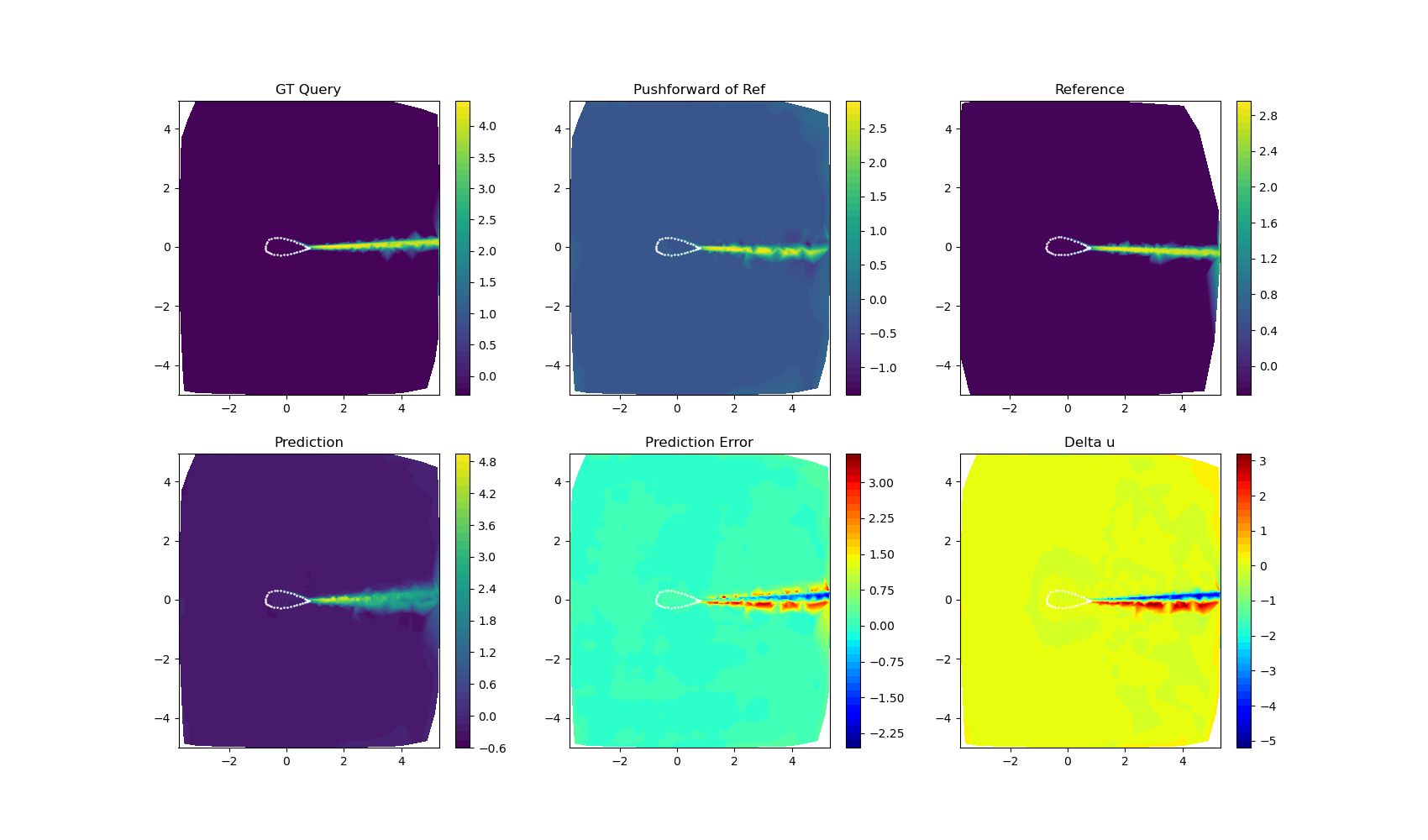}
    \caption{Qualitative example of RNO on the turbulent viscosity of Airfoil-RANS.}
    \label{fig:airf_rans}
\end{figure}

\textcolor{blue}{
Baseline models are GNOT and its modified version R-GNOT which adds reference as input. Note that unlike Airfoil-Euler where the mesh grids of every case is deformed from a standard uniform grids, many aforementioned baselines can not be easily implemented here since no such standard grids exists and the number of mesh grids are different from case to case. All metric is $l_2$. RNO achieves competitive results. See Table \ref{tab:airfoil_rans}. We note that, since AirfRANS was not designed to generate data for RNO, potentially RNO can improve its performance. We recommend generating datasets of reference and query pairs with \textit{fixed} boundary conditions (BCs) within a pair (BCs can be different between pairs) and \textit{varying} geometry, such that RNO can learn the change of solution due to the change of variable that we are interested in. Such setting can be particularly useful for e.g. sensitivity and robustness analysis. See qualitative examples in Fig. \ref{fig:airf_rans}.
}

\begin{table}[h!]
    \centering
    \begin{tabular}{c|cccc}
        \hline
         Models & $u$ & $v$ & $p$ & $\nu_t$ \\
         \hline
         GNOT & 0.51 & 0.44 & 0.61 & 0.31\\
         R-GNOT & \textbf{0.30} & \textbf{0.19} & \textit{0.11} & \textbf{0.19} \\
         RNO & \textit{0.32} & \textit{0.32} & \textbf{0.10} & \textit{0.22}\\
         \hline
    \end{tabular}
    \caption{\textbf{Bold} indicates the best and \textit{Italic} the second. $u,v$ are velocity components, $p$ the pressure, and $\nu_t$ the turbulent viscosity.}
    \label{tab:airfoil_rans}
\end{table}

\section{Distance-Aware Linear Attention}\label{app:linearAT}
To construct a linear attention for \eqref{crossAT}, we implement random feature mapping (RFM) \cite{rahimi2007random} to approximate the distance weighing function $\exp(-\|\cdot\|^2/\gamma^2)$. The same trick is applied by \cite{peng2021random} with a different purpose, to modify the softmax function in attention mechanism and achieve linear complexity\footnote{We found a typo in the paper of \cite{peng2021random}. In theorem 1. $\bm{w}_i$ should be sampled from $\mathcal{N}(\bm{0}, \sigma^2 \bm{I}_d)$, and then $\mathbb{E}_{\bm{w}_i}[\phi(\bm{x})\cdot\phi(\bm{y})]=\exp(-\|x-y\|^2\cdot\sigma^2/2)$, instead of $\exp(-\|x-y\|^2/2\sigma^2)$}. 

Let linear attention be $\textbf{attn}(x, y_j) = \frac{1}{N_{div}}\bm{q}(x)\cdot \bm{k}(y_j)$ where $N_{div}=\sum_j \bm{q}\cdot\bm{k}_j$ is a normalizing constant (see details of linear attention in \cite{hao2023gnot}), and \eqref{crossAT} becomes
\begin{align}
    \bm{w}(x) \approx \sum^N_j \sum^M_t \bm{q}(x)\cdot \bm{k}(y_j) \cdot h_{\gamma}(d(x, y_j)) \cdot \bm{v}^t(y_j) ,
\end{align}
To approximate $h_{\gamma}(d(x, y))=\exp(-\|x-y\|^2/\gamma^2)$, sample $\bm{\omega}_i$ from $\mathcal{N}(\bm{0}, \sigma^2 \bm{I}_d)$ where $\sigma=\frac{\sqrt{2}}{\gamma}$, and then define $\phi: \mathbb{R}^n\rightarrow\mathbb{R}^{2D}$ as
\begin{align}
    \phi(x) = \frac{1}{\sqrt{D}} \left[\sin(\bm{\omega}_1\cdot x),\cdots,\sin(\bm{\omega}_D\cdot x), \cos(\bm{\omega}_1\cdot x),\cdots,\cos(\bm{\omega}_D\cdot x)
    \right]
\end{align}
Then according to RFM we have
$\mathbb{E}_{\bm{\omega}_i}[\phi(x)\cdot\phi(y)]=\exp(-\|x-y\|^2/\gamma^2)$.
Let $\bm{q}, \bm{k}, \bm{v}, \phi(x)$ be indexed by $q_{ab}, k_{ab}, v_{ab}, \phi_{ac}$, where $a=1,\cdots,N$, $b=1,\cdots,d$, $c=1,\cdots,D$. Then let $j, m$ correspond to the first and the second indices $a, b$, and $l$ corresponds to $c$, the second index of $\phi$. We have
\begin{align*}
    w_{ab} &= \sum_{m,l,j} q_{am}k_{jm}\phi_{al}\phi_{jl}v_{jb} \\
            &= \sum_{m,l}  q_{am}\phi_{al} \sum_{j} k_{jm}\phi_{jl}v_{jb} \\
            &= \sum_{m} q_{am} \sum_l \phi_{al} H_{mlb} \\
            &= \sum_{m} q_{am} H_{amb},
\end{align*}
where $H_{mlb}=\sum_{j} k_{jm}\phi_{jl}v_{jb}$ and $H_{amb}=\sum_l \phi_{al} H_{mlb}$. The computational cost of $H_{mlb}$, $H_{amb}$ is $O(Nd^2D)$ and $O(d^2D)$. For all $w_{ab}$, $1\leq a\leq N$ and $1\leq b\leq d$, the cost is $O(Nd^2D)$. 

The performance of linear RNO is summarized in Table \ref{tab:linear_RNO},
\begin{table}[h]
    \centering
    \begin{tabular}{c|ccc|ccc}
        \hline
         &  \multicolumn{3}{|c|}{NS2d-360}  & \multicolumn{3}{|c}{NS2d-1440} \\
         & $u$ & $v$ & $p$ & $u$ & $v$ & $p$ \\
         \hline
        RNO-L & 4.2e-2 & 1.1e-1 & 4.1e-2 & 2.5e-2 & 6.6e-2 & 2.6e-2 \\
        RNO & 3.9e-2 & 1.1e-1 & 4.2e-2 & 1.9e-2 & 5.1e-2 & 2.3e-2 \\
         \hline
    \end{tabular}
    \caption{RNO-L stands for linear RNO and is comparable with RNO in accuracy.}
    \label{tab:linear_RNO}
\end{table}

RNO-L has a little higher error most likely due to the randomness introduced by RFM. We point out that a drawback of linear DACA is $O(d^2D)$ memory cost. When $D$ is taken a high integer value to improve the accuracy of estimation of the distance weight function $h_{\gamma}(d)$, the memory cost grows by $D$ times compared to vanilla linear attention. However, it can be mitigated by increasing number of heads of attention, with number of heads $n_{head}$, the memory cost is reduced to $O(\frac{d^2D}{n_{head}})$.

\section{Learning Objective}\label{app:model_details}
Consider a reference solution $u_r$ on $\Omega_r$ and a query domain $\Omega_q$. Given a vector field $V=V(x,t)$ and the corresponding flow $T_t$, recall the material derivative \cite{sokolowski1992introduction} of the solution operator $G$ at $\Omega_r$, 
\begin{align}
    \dot u(\Omega_r, V)=\lim_{t\rightarrow 0} \frac{1}{t} (u_{T_t(\Omega_r)}\circ T_t - u_{\Omega_r}),
\end{align}
where $\circ$ is composition of functions and $u_{T_t(\Omega_r)}\circ T_t$ is the pullback of $u_{T_t(\Omega_r)}$. The limit exists in a suitable topology, for example a Sobolev space $W^{m,p}(\Omega)$ or $H^1_0(\Omega)$ \cite{evans2022partial}.
Suppose $t=\varepsilon$ and $T_{\varepsilon}=\varphi$, then we can rewrite the above equation as
\begin{align}
    u_{\Omega_q}\circ \varphi = u_{\Omega_r} + \varepsilon \cdot \dot u(\Omega_r, V) + o(\varepsilon).
\end{align}
where $u_{\Omega_q}=u_q$. 
Suppose $\varepsilon \cdot \dot u(\Omega_r, V)$ is approximated by a neural operator $\Psi_{\theta}$. 
Then for $x\in\Omega_r$,
\begin{equation}
    u_q\circ\varphi(x) \approx u_r(x) + \Psi_{\theta}\left((x, u_r), \Omega_r, \varphi\right).
\end{equation} 
For a set of points $\{x_{q_i}\}_i\subset\Omega_q$, and a collection of $\Omega_q$'s of size $N$, each paired with a reference triplet $(u_r, \Omega_r, \varphi)$, the objective function \eqref{objective0} is approximated by
    \begin{dmath}
    \label{objective1}
    \min_{\theta}  \frac{1}{N}\sum_q\sum_i \|\Psi_{\theta}(u_r, \Omega_r, \varphi)(x_{r_i}) - (u_q\circ \varphi - u_r)(x_{r_i})\|.
\end{dmath}
Note that this objective is computed on the domain $\Omega_r$. Alternatively, we can transform the problem to $\Omega_q$. Let $\{x_{r_i}\}_i\subset\Omega_r$ and $\varphi(x_{r_i})=x_{q_i}\in\Omega_q$ for all $i=0,1,2\ldots$ The above objective can be rewritten as
\begin{dmath}\label{objective2}
    \min_{\theta}  \frac{1}{N}\sum_q\sum_i \|\Psi_{\theta}(u_r, \Omega_r, \varphi)(\varphi^{-1}(x_{q_i})) - (u_q - u_r\circ\varphi^{-1})(x_{q_i})\|.
\end{dmath}
Let $\delta u = \Psi_{\theta}(u_r, \Omega_r, \varphi)\circ\varphi^{-1} $, and then predicted solution $\hat{u}_q = u_r\circ\varphi^{-1} + \delta u$. Given ground truth $u_q$, the above objective is implemented as the metric loss between $\hat{u}_q$ and $u_q$.

A minor difference between \eqref{objective1} and \eqref{objective2} is due to interpolation. Namely, $u_q\circ \varphi(x_{r_i})$ can be approximated by interpolating $u_q$ on $\varphi(x_{r_i})$ in $\Omega_q$, and $u_r\circ\varphi^{-1}(x_{q_i})$ can be approximated by interpolating $u_r$ on $\varphi^{-1}(x_{q_i})$ in $\Omega_r$. In our implementation, we adopt \eqref{objective} to leave the target $u_q$ unchanged, but the other way is obviously also valid. 


\section{Error Analysis}\label{app:error}
We compare a natural baseline, the pushforward of reference solution, $u_r\circ\varphi^{-1}$, with RNO. See Table \ref{tab:pushforward}. \textcolor{blue}{It shows that RNO shrinks the gap between $u_r\circ\varphi^{-1}$ and target $u_q$. In fact, what RNO is actually learning is to fill this gap.}
\begin{table}[h]
    \centering
    \begin{tabular}{c|cccc}
        \hline
         & NS2d & NS2d-sq & Inductor2d & Heatsink3d \\
         \hline
        $u_r\circ\varphi^{-1}$ & 1.2e-1 & 2.5e-1 & 7.3e-2 & 1.6e-1\\
        RNO & \textbf{6.3e-2} & \textbf{1.1e-1} & \textbf{1.5e-2} & \textbf{8.6e-2} \\
        \hline
    \end{tabular}
    \caption{Mean error of the pushforward reference solution and RNO.}
    \label{tab:pushforward}
\end{table}

\begin{figure}[h]
    \centering
    \includegraphics[width=\linewidth]{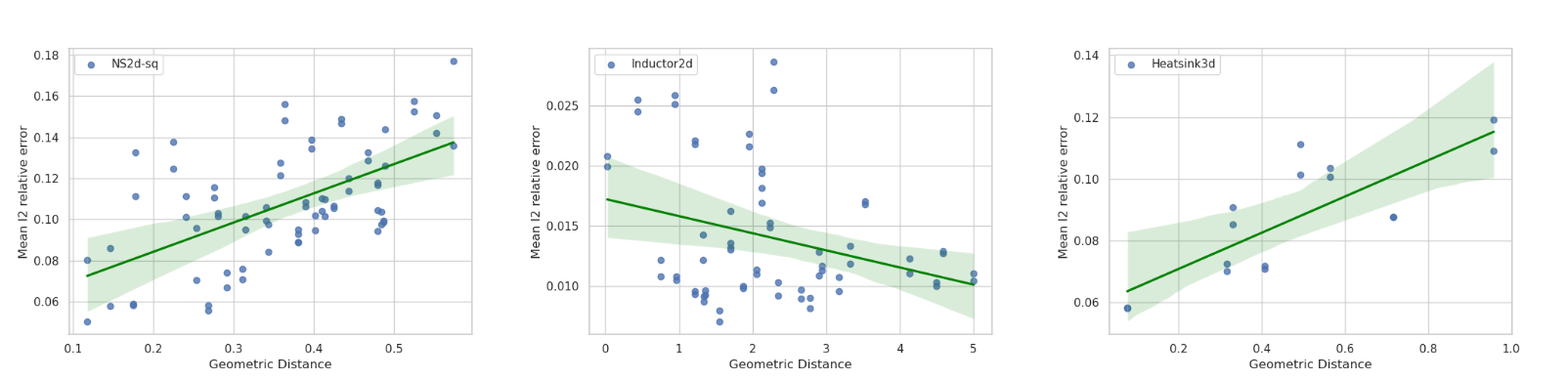}
    \caption{Left to right: Error versus distance of NS2d-sq, Inductor2d, Heatsink3d.}
    \label{fig:error_dist}
\end{figure}

In Fig. \ref{fig:error_dist}, we plot mean $l_2$ relative error versus geometric distance for other datasets besides NS2d. Error grows as distance increases except for Inductor2d. Our speculation is that geometric distance defined by the Euclidean distance between parameters of reference and query may not accurately measure deformations in this case. \textcolor{blue}{A more accurate alternative for geometric distance can be the norm of deformation $\varphi$ from reference to query geometry.}

\end{document}